\pdfoutput=1

\documentclass[11pt]{article}

\usepackage{ACL2023}
\usepackage{adjustbox}
\usepackage{times}
\usepackage{xcolor}
\usepackage{float}
\usepackage{latexsym}
\usepackage{hhline}
\usepackage{multirow}
\usepackage{dcolumn}
\newcolumntype{d}[1]{D{.}{.}{#1}}
\newcolumntype{t}[1]{D{:}{:}{#1}}
\usepackage{makecell}
\usepackage{pdflscape}

\newcommand{\errorCells}{\multicolumn{1}{c}{NL} & \multicolumn{1}{c}{WL} & \multicolumn{1}{c}{UN} & \multicolumn{1}{c}{X}  & \multicolumn{1}{c|}{E}}
\newcommand{\CorrectCells}{\multicolumn{1}{c}{CS} & \multicolumn{1}{c}{CB} & \multicolumn{1}{c}{CN} & \multicolumn{1}{c}{CC} & \multicolumn{1}{c||}{C}}
\newcommand{\CorrectCellsEnd}{\multicolumn{1}{c}{CS} & \multicolumn{1}{c}{CB} & \multicolumn{1}{c}{CN} & \multicolumn{1}{c}{CC} & \multicolumn{1}{c}{C}}

\newcommand{\EvaluationMetric}{\multicolumn{1}{c}{Chrf} & \multicolumn{1}{c}{Chrf++} & \multicolumn{1}{c}{spBLEU} & \multicolumn{1}{c|}{BLEU} }
\newcommand{\EvaluationMetricEnd}{\multicolumn{1}{c}{Chrf} & \multicolumn{1}{c}{Chrf++} & \multicolumn{1}{c}{spBLEU} & \multicolumn{1}{c}{BLEU} }

\newcommand{\CorrelationMetric}{\multicolumn{1}{c}{Pearson-corr} & \multicolumn{1}{c|}{p-value} }
\newcommand{\CorrelationMetricEnd}{\multicolumn{1}{c}{Pearson-corr} & \multicolumn{1}{c}{p-value} }
\usepackage[T1]{fontenc}

\usepackage{lipsum}

\usepackage{microtype}

\usepackage{inconsolata}

\usepackage{footnote}
\usepackage{tablefootnote}
\makesavenoteenv{tabular}

\newcommand{\footURL}[1]{\footnote{\url{#1}}}

\makeatletter
\newcommand\footnoteref[1]{\protected@xdef\@thefnmark{\ref{#1}}\@footnotemark}
\makeatother

\usepackage{ragged2e}
\usepackage{tabularx}
\newcolumntype{Y}{>{\centering\arraybackslash}X}
\newcolumntype{Z}{>{\raggedleft\arraybackslash}X}

\usepackage{hyperref}
\hypersetup{
colorlinks   = true,
citecolor    = blue,
urlcolor    =  blue
}
\usepackage{url}

\usepackage{soul}
\newcommand*{\affmark}[1][*]{\textsuperscript{#1}}

\newcommand{\Si}{\texttt{Si}}
\newcommand{\En}{\texttt{En}}
\newcommand{\Ta}{\texttt{Ta}}
\newcommand{\EnSi}{\texttt{En-Si}}
\newcommand{\EnTa}{\texttt{En-Ta}}
\newcommand{\SiTa}{\texttt{Si-Ta}}

\title{Quality Does Matter: A Detailed Look at the Quality and Utility of Web-Mined Parallel Corpora}

\author{Surangika Ranathunga\affmark[1], Nisansa de Silva\affmark[2], Menan Velayuthan\affmark[2],\\ \textbf{Aloka Fernando\affmark[2], Charitha Rathnayake\affmark[2]} \\
 \affmark[1]Massey University, Palmerston North, New Zealand, 4443 \\
 \affmark[2]Dept. of Computer Science \& Engineering, University of Moratuwa,10400, Sri Lanka \\
  \texttt{s.ranathunga@massey.ac.nz} \\
  \texttt{\{NisansaDdS,velayuthan.22,alokaf,charitha.18\}@cse.mrt.ac.lk}  
}

\begin{document}
\maketitle
\begin{abstract}
We conducted a detailed analysis on the quality of web-mined corpora for two low-resource languages (making three language pairs, English-Sinhala, English-Tamil and Sinhala-Tamil). We ranked each corpus according to a similarity measure and carried out an intrinsic and extrinsic evaluation on different portions of this ranked corpus. We show that there are significant quality differences between different portions of web-mined corpora and that the quality varies across languages and datasets. We also show that, for some web-mined datasets, Neural Machine Translation (NMT) models trained with their highest-ranked 25k portion can be on par with human-curated datasets.
\end{abstract}

\section{Introduction}
Despite the advances in NMT research, the availability of parallel corpora is still a deciding factor of NMT model performance. This puts low-resource languages at a clear disadvantage~\cite{ranathunga2023neural}. Even the use of Pre-trained Language Models (PLMs) is not quite enough to overcome the impact of data scarcity~\cite{lee-etal-2022-pre}. 

Publicly available web-mined parallel corpora (bitext) such as CCMatrix~\citep{schwenk-etal-2021-ccmatrix}, CCAlign~\citep{el-kishky-etal-2020-ccaligned}, WikiMatrix~\citep{schwenk-etal-2021-wikimatrix}, NLLB~\citep{nllb2022}, and ParaCrawl~\cite{banon-etal-2020-paracrawl} bring a glimmer of hope against this data scarcity problem. Compared to human-curated datasets, these are larger in quantity and contain data for hundreds of languages, including several low-resource languages. There are further initiatives to mine bitext for yet more languages as well~\cite{bapna2022building}.

However, \citet{kreutzer-etal-2022-quality} analysed a sample of 100 sentence pairs from some of these corpora and showed that these web-mined corpora have serious quality issues, especially for low-resource languages.~\citet{lee-etal-2022-pre} noticed a drop in NMT results when a model was trained using a random 100k  sample of CCAlign.~\citet{khayrallah-koehn-2018-impact}   injected different noise types found in web-mined corpora (by analysing a random sample) into a clean parallel corpus and showed that it has a debilitating impact on NMT performance. 

These findings paint a grim picture of the utility of web-mined corpora. However, they all considered a random sample of these corpora to determine their quality. This implicitly assumes that the quality 
is consistent throughout the corpus.

In this research, we show that analysing a random sample of such large web-mined corpora can be misleading. We selected parallel corpora for two low-resource languages Sinhala and Tamil, which made three language pairs pairs: English-Sinhala (\EnSi), English-Tamil (\EnTa) and Sinhala-Tamil (\SiTa). Instead of quality checking a very small random sample of a web-mined corpus as done by~\citet{kreutzer-etal-2022-quality}, we ranked the sentence pairs by means of a similarity measure and extracted top 25k, bottom 25k and a random 25k portions of each corpus. 

We improved the error taxonomy of~\citet{kreutzer-etal-2022-quality} and carried out a human (intrinsic) evaluation on a random sample of 250 from each of these portions. Our results show that there are significant quality differences between the three portions, and the quality of the top 25k portion is much better than the other portions. We also noted major variations of quality across web-mined corpora belonging to different language pairs. 

We then carried out an extrinsic evaluation. We separately trained NMT systems by using these top, bottom, as well as the random 25k samples of the corpora and tested them with two different evaluation sets. These results also showed that NMT models trained with the top 25k portion are significantly better. NMT models trained with the full version of some of these corpora were even lagging behind models trained with their top 25k portion. The NMT model trained with the top 25k portion of the \EnSi{} and \EnTa{} parts of the NLLB corpus performed even better than a model trained with a human-curated corpus.

We then fixed the translation issues in the top 25k of the NLLB corpus using human translators. The time taken to clean the corpus was slightly less than the time taken to translate the corpus from scratch. Although an NMT model trained with this cleaned corpus outperformed the uncleaned corpus, the resultant meagre gains cannot be justified when considering the time and money spent on the translators.

In summary, our results caution the researchers not to haphazardly use the web-mined corpora with just random sampling. Simply ranking a web-mined corpus first and then using only the high-quality portion would result in better accuracy in much less training time.  We also hope other researchers (especially those working on low-resource languages) would carry out similar analyses for datasets of their languages. This will help future researchers make informed decisions when selecting web-mined corpora for NMT research.

\section{Related Work}
\label{sec:related_work}
Web-mined parallel corpora are gathered from any available website without guarantees about quality. \citet{khayrallah-koehn-2018-impact, lee-etal-2022-pre} pointed out that NMT systems built with such web-mined corpora have performance issues. 

The common way to determine the quality of a parallel corpus is by analysing the performance of a Machine Translation system trained with that corpus~\cite{khayrallah-koehn-2018-impact, schwenk-etal-2021-wikimatrix, koehn-etal-2020-findings}. However, this does not indicate the types of noise in the corpus.

Human evaluation of the quality of parallel sentences (let them be web-mined, machine-generated, or human-generated) requires some criteria for the evaluators to make a judgement. \citet{bojar-etal-2016-findings} introduced the \textit{Direct Assessment} criteria, where each sentence pair is ranked on a 0-100 scale. However, such a numerical scale does not shed light on the different types of noise in web-mined corpora. 

\citet{khayrallah-koehn-2018-impact} analysed a web-mined corpus and introduced the first categorisation of noise. The categories are: \textit{misaligned sentences}, \textit{mis-ordered words}, \textit{wrong language}, \textit{untranslated sentences}, and \textit{short segments}. \citet{herold-etal-2022-detecting} extended this categorisation with three new classes: \textit{raw crawled data}, \textit{over/under-translation}, and \textit{synthetic translation}.

\begin{table}[!htb]
    \centering
    \resizebox{0.49\textwidth}{!}{%
    \begin{tabular}{lrrrrr}
    \hline
                       & CCAligned & Wikimatrix & CCMatrix & XLEnt & NLLB \\
    \hline
      \textbf{\EnSi}   & 619,729 & 115,045 & 6,270,800 & 690,186 & 24,336,367 \\
      \textbf{\EnTa}   & 878,689 & 95,161 & 7,291,118 & 634,299 & 42,588,178 \\
      \textbf{\SiTa}   & -       & -      & 215,965 & 153,532 & 1,493,318 \\
      Source           & Common Crawl
      & Wikipedia &    Common Crawl  &         &            \\
      Filtering Level & document & sentence &   sentence     &         &            \\
      Alignment       & LASER & LASER &   LASER      &   LASER      &    LASER-3     \\
    \hline
    \end{tabular}
    }
    \caption{Dataset Statistics}
\label{tab:data_stats}
\end{table}

In contrast to the above categorisations,~\citet{kreutzer-etal-2022-quality}'s taxonomy has labels for both correct and erroneous sentence pairs: 1.) Correct translation - natural sentence, 2.) Correct translation but Boilerplate or low quality, 3.) Correct translation - short, 4.) Incorrect translation but both correct languages, 5.) Source OR target wrong language but both still linguistic content, and 6.) Not a language.

\citet{kreutzer-etal-2022-quality} conducted a  human evaluation using their taxonomy for three web-mined corpora (CCAligned, ParaCrawl v7.1, WikiMatrix) and covered data from both high and low resource languages. \citet{de-gibert-bonet-etal-2022-quality} used that taxonomy to evaluate English-Catalan corpora.

\section{Languages}
We selected three language pairs: English-Sinhala (\EnSi), English-Tamil (\EnTa) and Sinhala-Tamil (\SiTa). Tamil (\Ta) and Sinhala (\Si) are large institutional languages~\cite{ethnologue}. However, considering their data availability,~\citet{joshi-etal-2020-state} categorised Tamil as a mid-resource language and Sinhala as an extremely low-resource language. In the more recent language categorization by~\citet{ranathunga-de-silva-2022-languages}, Sinhala has moved one class up, and the position of Tamil is unchanged. Sinhala, in particular, is contained only in the island nation of Sri Lanka, and has only seen slow progress in language technologies~\cite{ranathunga-de-silva-2022-languages,de2023survey}. But, being a multilingual country, translation systems are of utmost importance to Sri Lanka. This is particularly true for \SiTa, as most government documents are first prepared in Sinhala and then translated to Tamil and English~\cite{farhath2018integration}.

\begin{table*}[!htb]
\centering
\includegraphics[width=\textwidth]{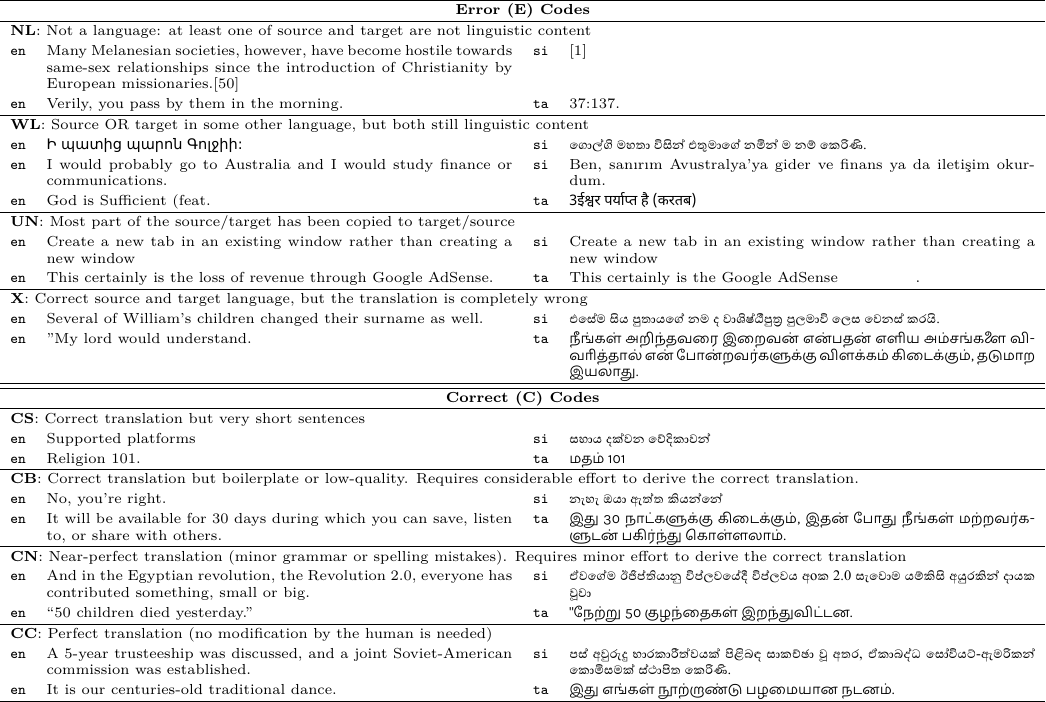}
\caption{Quality Evaluation Taxonomy with En-Si \& En-Ta examples}
\label{tab:taxonomy}
\end{table*}

\section{Web-mined Parallel Corpora}

Table~\ref{tab:data_stats} lists the web-mined corpora that we considered for evaluation. Other web-mined corpora available in OPUS~\cite{tiedemann-2012-parallel} were omitted because they did not have at least 100k samples for at least two of the language pairs we considered. Out of the selected corpora, XLEnt~\citep{el-kishky-etal-2021-xlent} was later omitted from human evaluation, because it has a significant amount of single-words or short phrases in its top 25k portion. These corpora are further described in Appendix~\ref{app:corpora}.


\section{Quality Estimation by Humans}
\label{sec:quality_est_human}

As mentioned earlier,~\citet{kreutzer-etal-2022-quality} carried out the first human evaluation on the quality of web-mined corpora. Although they reported results for a large number of languages including low-resource languages, their discussion was mainly centred around the language-wise aggregated results. Thus they only used randomly selected 100 sentences from each language-specific corpus. \citet{de-gibert-bonet-etal-2022-quality} carried out a similar study for Catalan-English, but they also considered only 100 samples from a corpus.

In contrast, we carried out a more detailed analysis of web-mined corpora belonging to the three language pairs by first ordering each parallel corpus according to the quality of the sentence pair. Our hypothesis is that the quality of a web-mined corpus is not consistent across a dataset, thus analysing a random portion of the corpus would not give a clear picture of the quality of the corpus.

\begin{table}[!htb]
\small
\begin{tabular}{| p{0.1\linewidth} | p{0.8\linewidth} |}
\hline
Ours & \citet{herold-etal-2022-detecting}'s \\
\hline
NL &  \\
\hline
WL & Wrong Language (src|trg) \\
\hline
UN & Untranslated (src|trg) \\
\hline
X & Misaligned Sentences \\
\hline
CS & Short Segments (max. length) \\
\hline
{CB} & Misordered Words (src|trg),
Raw Crawled Data,
Over-/Under translation,
Synthetic Translations \\
\hline
CN &  \\
\hline
CC &  \\
\hline

\end{tabular}
\caption{Comparison of our taxonomy with the error Categories in~\citet{herold-etal-2022-detecting}}
\label{tab:herold}
\end{table}

\begin{table*}[!htb]
    \centering
\resizebox{\textwidth}{!}{%
    \begin{tabular}{ll|*{5}{d{2.2}}|*{5}{d{2.2}}||*{5}{d{2.2}}|*{5}{d{2.2}}||*{5}{d{2.2}}|*{5}{d{2.2}}}
        \hline
        \multicolumn{2}{l|}{\multirow{2}{*}{Dataset}} & \multicolumn{10}{c||}{\EnSi}  &   \multicolumn{10}{c||}{\EnTa} & \multicolumn{10}{c}{\SiTa}  \\
        \hhline{~~------------------------------}
        & & \errorCells & \CorrectCells & \errorCells & \CorrectCells & \errorCells & \CorrectCellsEnd \\
        \hline
        \multirow{3}{*}{CCAligned} & Top & 0.0 & 0.0 & 1.9 & 0.3 & 2.2 & 13.2 & 59.7 & 10.5 & 14.4 & 97.8 & 0.1 & 0.1 & 5.5 & 0.4 & 6.1 & 18.7 & 33.6 & 25.3 & 16.3 & 93.9 & - & - & - & - & - & - & -& -& - & - \\ 
         & Random & 2.0 & 0.1 & 5.9 & 8.9 & 16.9 & 17.9 & 36.1 & 13.2 & 15.9 & 83.1 & 0.4 & 0.0 & 0.9 & 25.9 & 27.2 & 9.1 & 28.1 & 19.9 & 15.7 & 72.8 & - & - & - & - & - & - & -& -& - & - \\
        & Bottom & 0.5 & 0.0 & 0.1 & 60.4 & 61.0 & 4.3 & 17.5 & 11.3 & 5.9 & 39.0 & 1.9 & 0.1 & 1.1 & 45.5 & 48.6 & 8.8 & 10.0 & 16.1 & 16.5 & 54.1 & - & - & - & - & - & - & -& -& - & - \\        
        \hline

        \multirow{3}{*}{WikiMatrix} & Top &  0.3& 0.1 & 2.1 & 16.3 & 18.8 & 6.1 & 40.8 & 12.0 & 22.3 & 81.2 & 0.9 & 6.3 & 15.9 & 46.1 & 69.2 & 1.6 & 10.3 & 7.5 & 11.5 & 30.9 & - & - & - & - & - & - & -& -& - & - \\ 
        & Random & 0.3 & 0.1 & 0.0 & 86.1 & 86.5 & 1.2 & 7.9 & 2.9 & 1.5 & 13.5 & 0.7 & 0.9 & 1.2 & 91.9 & 94.7 & 0.3 & 4.1 & 0.7 & 0.3 & 5.4 & - & - & - & - & - & - & -& -& - & - \\
        & Bottom & 0.0 & 2.7 & 0.3 & 88.5 & 91.5 & 1.2 & 6.9 & 0.4 & 0.0 & 8.5 & 1.3 & 5.2 & 0.8 & 88.7 & 96.0 & 0.3 & 2.1 & 1.2 & 0.4 & 4.0 & - & - & - & - & - & - & -& -& - & - \\         
        \hline

        \multirow{3}{*}{CCMatrix} & Top & 0.0 & 0.0 & 7.1 & 0.1 & 7.2 & 8.7& 37.5 & 14.3 & 32.4 & 92.9 & 0.0 & 0.0 & 51.5 & 3.6 & 55.1 & 2.7 & 27.5 & 8.5 & 6.3 & 45.0 & 0.1 & 5.5 & 0.5 & 2.1 & 8.2 & 9.3 & 26.4 & 34.3 & 21.7 & 91.7 \\
        & Random & 0.0 & 0.0 & 1.6 & 31.3 & 32.9 & 6.1 & 27.6 & 22.5 & 10.8 & 67.0 & 0.1 & 0.0 & 2.9 & 83.5 & 86.5 & 0.3 & 8.5 & 3.1 & 1.6 & 13.5 & 0.0 & 2.1 & 0.8 & 31.3 & 34.2 & 0.9 & 34.7 & 23.2 & 6.9 & 65.7\\
        & Bottom & 0.0 & 1.3 & 0.8 & 27.2 & 29.3 & 7.3 & 47.3 & 8.5 & 7.5 & 70.7 & 0.0 & 0.1 & 0.0 & 83.1 & 83.2 & 0.0 & 10.1 & 4.3 & 2.4 & 16.8 & 0.0 & 1.2 & 0.1 & 50.1 & 51.4 & 1.1 & 31.7 & 11.3 & 4.4 & 48.6 \\
        
        \hline

       \multirow{3}{*}{NLLB} & Top & 0.0 & 0.5 & 0.4 & 19.1 & 20.0 & 6.5 & 36.0 & 18.8 & 18.7 & 80.0 & 0.0 & 0.4 & 0.3 & 11.1 & 11.8 & 0.4 & 21.6 & 25.2 & 41.1 & 88.3 & 0.0 & 0.0 & 0.3 & 1.9 & 2.2 & 0.3 & 22.3 & 40.5 & 34.8 & 97.9 \\
       & Random & 0.1 & 0.4 & 0.7 & 54.5 & 55.7 & 1.3 & 27.5 & 10.5 & 4.9 & 44.2 & 0.1 & 0.0 & 0.5 & 43.3 & 1.9 & 43.9 & 31.6 & 10.9 & 11.6 & 98.0 & 56.0 & 0.3 & 0.0 & 20.0 & 20.3 & 1.2 & 44.5 & 22.3 & 11.7 & 79.7 \\ 
        & Bottom & 0.0 & 0.0 & 1.9 & 56.9 & 58.8 & 4.7 & 27.1 & 8.1 & 1.3 & 41.2 & 0.0 & 0.0 & 0.0 & 51.9 & 51.9 & 1.6 & 28.9 & 11.1 & 6.5 & 48.1 & 0.0 & 0.0 & 0.1 & 34.7 & 34.8 & 0.0 & 42.0 & 20.3 & 2.9 & 65.2 \\
        \hline
       \multirow{3}{*}{NLLB (cleaned)} & Translator 1 & 0.0 & 0.0 & 0.0 & 1.9 & 1.9 & 10.7 & 24.0 & 14.3 & 49.1 & 98.1 & 0.1 & 0.0 & 0.0 & 0.0 & 0.1 & 0.5 & 16.2 & 12.6 & 70.6 & 99.9 & 0.0 & 0.0 & 0.0 & 0.3 & 0.3 & 0.3 & 1.9 & 24.0 & 73.6 & 99.7\\
        & Translator 2 & 0.0 & 0.1 & 0.0 & 1.9 & 2.0 & 7.2 & 21.3 & 13.0 & 55.6 & 98.0 & 0.0 & 0.0 & 0.0 & 0.1 & 0.1 & 0.8 & 16.9 & 10.1 & 72.1 & 99.9 & 0.0 & 0.0 & 0.0 & 0.4 & 0.4 & 0.3 & 4.3 & 38.0 & 57.0 & 99.6\\
        & Translator 3 & 0.0 & 0.4 & 0.0 & 1.8 & 2.1 & 9.1 & 22.5 & 7.0 & 59.3 & 97.9 & 0.0 & 0.0 & 0.1 & 0.4 & 0.5 & 0.6 & 8.1 & 15.8 & 75.0 & 99.5 & 0.0 & 0.0 & 0.0 & 0.0 & 0.0 & 0.1 & 1.9 & 29.5 & 68.5 & 100.0\\
        \hline

    \end{tabular}
    }
    \caption{The average percentage of tag counts over 3 independent evaluators for \EnSi, \EnTa, and \SiTa~for 250 samples from top, bottom and random splits. \texttt{C} - sum of \texttt{CS}, \texttt{CB}, \texttt{CN} and \texttt{CC}. \texttt{E} - sum of \texttt{NL}, \texttt{WL}, \texttt{UN}, and \texttt{X}.}    \label{tab:human_eval_results}
\end{table*}

\paragraph{Participants:}Fifteen translators were employed to conduct the human evaluation across the three language pairs. Evaluator selection and training details are in Appendix~\ref{app:appendix_human_details}. 

\paragraph{Sample Selection:}Calculating a similarity measure over the source and target sentence embeddings is a popular method to get an indication of the quality of a parallel sentence pair~\cite{koehn-etal-2020-findings}. We picked LASER-3~\cite{heffernan-etal-2022-bitext} as our apparatus to score the alignment between the bitext.~\citet{heffernan-etal-2022-bitext} demonstrated that LASER-3 performs either on par or better than  LaBSE\footURL{https://tfhub.dev/google/LaBSE/2},
the other commonly used multilingual sentence encoder.

Sentences in each corpus were ordered by the LASER-3 score. For the NLLB corpus, we used the LASER-3 scores that were already provided within the dataset. For other datasets, we calculated this score\footURL{https://github.com/facebookresearch/LASER}.
From this sorted corpus, we randomly selected 250 sentences each from the top 25k split, the bottom 25k split, as well as from the entire corpus. There was no overlap between the sentences selected from the random set and the top/bottom sets. Once again, be reminded that~\citet{kreutzer-etal-2022-quality} used only 100 random sentences from the entire corpus.

\paragraph{Taxonomy:}Our error taxonomy shown in Table~\ref{tab:taxonomy} is based on~\citet{kreutzer-etal-2022-quality} and~\citet{herold-etal-2022-detecting}. 
Unlike~\citet{kreutzer-etal-2022-quality}, we manually cleaned a web-mined corpus to determine its effect on NMT performance (see Section~\ref{sec:impact_manual_cleaning}). Therefore our taxonomy indicates the level of human effort needed to fix the translation of a pair of sentences. We believe this provides more guidance to humans conducting quality evaluations of the corpora. 
Compared to~\citet{kreutzer-etal-2022-quality}, our taxonomy has two other differences: (1) We used \verb|WL| to denote when the source or target is in some third language, and \verb|UN| to denote when source or target has been copied to the other side. In contrast,~\citet{kreutzer-etal-2022-quality} used \verb|WL| to denote both of these scenarios. (2)~\citet{kreutzer-etal-2022-quality} used \verb|CC| to denote both perfect and near-perfect translations. In contrast, we used \verb|CC| only for perfect translations and introduced \verb|CN| for a near-perfect translation. While a bitext mining system may not be able to distinguish between \verb|CC| and \verb|CN|, this difference is important when manually cleaning the corpus. 

Comparison of our taxonomy against~\citet{herold-etal-2022-detecting} is given in Table~\ref{tab:herold}. Since they only focused on identifying errors, they do not have any category related to correct translation pairs. \citet{herold-etal-2022-detecting} used their error categories to introduce synthetic errors to a clean corpus. Therefore they could easily generate data that corresponds to  Mis-ordered Words (\texttt{src|trg}), Raw Crawled Data, Over/Under translation and Synthetic Translations. However, such errors are not directly distinguishable by a human. On the other hand, a sentence pair with at least one of these errors requires significant human effort to get cleaned. Therefore we grouped those categories as \verb|CB|.

\section{Human Evaluation Results}
\label{sec:human_results}

Each sentence pair was evaluated by three evaluators. The average agreement (measured in Pearson correlation) per language pair is as follows: \EnSi{} 0.40, \EnTa{} 0.55 and \SiTa{} 0.57 (Detailed results are in  Table~\ref{tab:correlation} of Appendix~\ref{app:appendix_human_details}). 
Results in Table~\ref{tab:human_eval_results} confirm 3 important points:

\paragraph{1.}The quality of a web-mined corpus is not consistent throughout. We see drastic differences in quality between the top 25k and the bottom 25k. For example, the top 250 samples of the \EnSi~WikiMatrix corpus have 34.3\% sentences falling into \verb|CC+CN| categories, while its bottom portion has only 0.4\% in the same categories. 

\paragraph{2.}Carrying out a human evaluation on a random sample as done by~\citet{kreutzer-etal-2022-quality} portrays a high amount of quality issues. For WikiMatrix, CCMatrix, and NLLB, random sampling gives results that are closer to the bottom than the top. CCAligned defies this trend strongly in \EnSi~and weakly in~\EnTa. 

\paragraph{3.} Quality of corpora can vary significantly depending on the language pair. For example, CCMatrix \EnSi{} top 25k has 46.7\% of \verb|CC+CN| categories, and the same for \EnTa~is 14.8\%.

Together, these observations warn us against haphazardly using these web-mined corpora without studying their quality distribution. 
The result for non-English-centric \SiTa{} is of particular interest. For \SiTa, both NLLB and CCMatrix top portion seem to be extremely good. In fact, the 97.9\% total value for the Correct (\verb|C|) group is the highest across all the results. 

~\citet{kreutzer-etal-2022-quality} did not consider \EnTa{} or \SiTa{} in their evaluation. Even for \EnSi, only the ParaCrawl v7.1 corpus was considered. Therefore we cannot draw a direct comparison with their results. However, we can compare their micro-averaged results with our results for the random split, for the same corpora. For the CCAligned corpus, our random split results for both \EnSi{} and \EnTa{} are significantly higher than \citet{kreutzer-etal-2022-quality}'s. In contrast, the same for WikiMatrix is lower than \citet{kreutzer-etal-2022-quality} by 10.24 and 18.34 (respectively).  Even though~\citet{kreutzer-etal-2022-quality} reported that $7.3\%$ of the languages they analyzed did not contain a single correct sentence, we observed a similar phenomenon only with the bottom 25k split of WikiMatrix. These observations further justify the need for language-specific detailed analysis of web-mined corpora. 

\section{Qualitative Analysis of Corpora}
\label{sec:qualitative_analysis}
In addition to the human evaluation discussed in the previous section, we also carried out a manual inspection of the top 25k portion of each corpus. 

\begin{table}[!htb]
\centering
\includegraphics[width=0.48\textwidth]{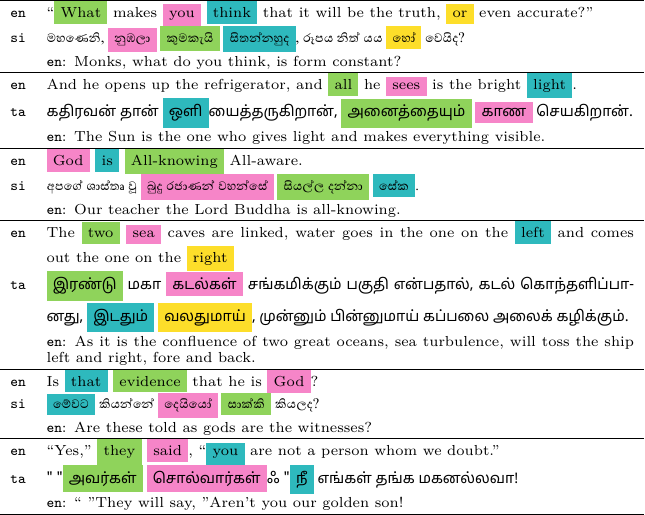}
\caption{Examples of \textit{parallel} sentences from NLLB where the translated \Si{} or \Ta{} sentence has a different meaning than the original \En{} sentences. We colour-coded the pairs of semantically close words that possibly contributed to the misalignment. Correct \En{} translation of the \Si{}/ \Ta{} sentence is also given for comparison. }
\label{tab:BadParallel}
\end{table}

Similar to~\citet{kreutzer-etal-2022-quality}, in \EnSi{} and \EnTa{} corpora, we found instances where sentences that are structurally and semantically similar but not parallel, presented as pairs. Table~\ref{tab:BadParallel} shows some such interesting examples from NLLB (An extended version is in Appendix~\ref{sec:badparallel}
 as Table~\ref{tab:BadParallelLong}). 
 We show instances where text from the \textit{Bible} has been aligned with \textit{Buddhist} scripture as well as instances where simple negation and noun matching have resulted in faulty alignments.
 \citet{kreutzer-etal-2022-quality} noted that such misaligned data may cause trained models to hallucinate fake \textit{facts}.

\begin{figure}[tb]
    \centering
    
    \includegraphics[width=0.4\textwidth]{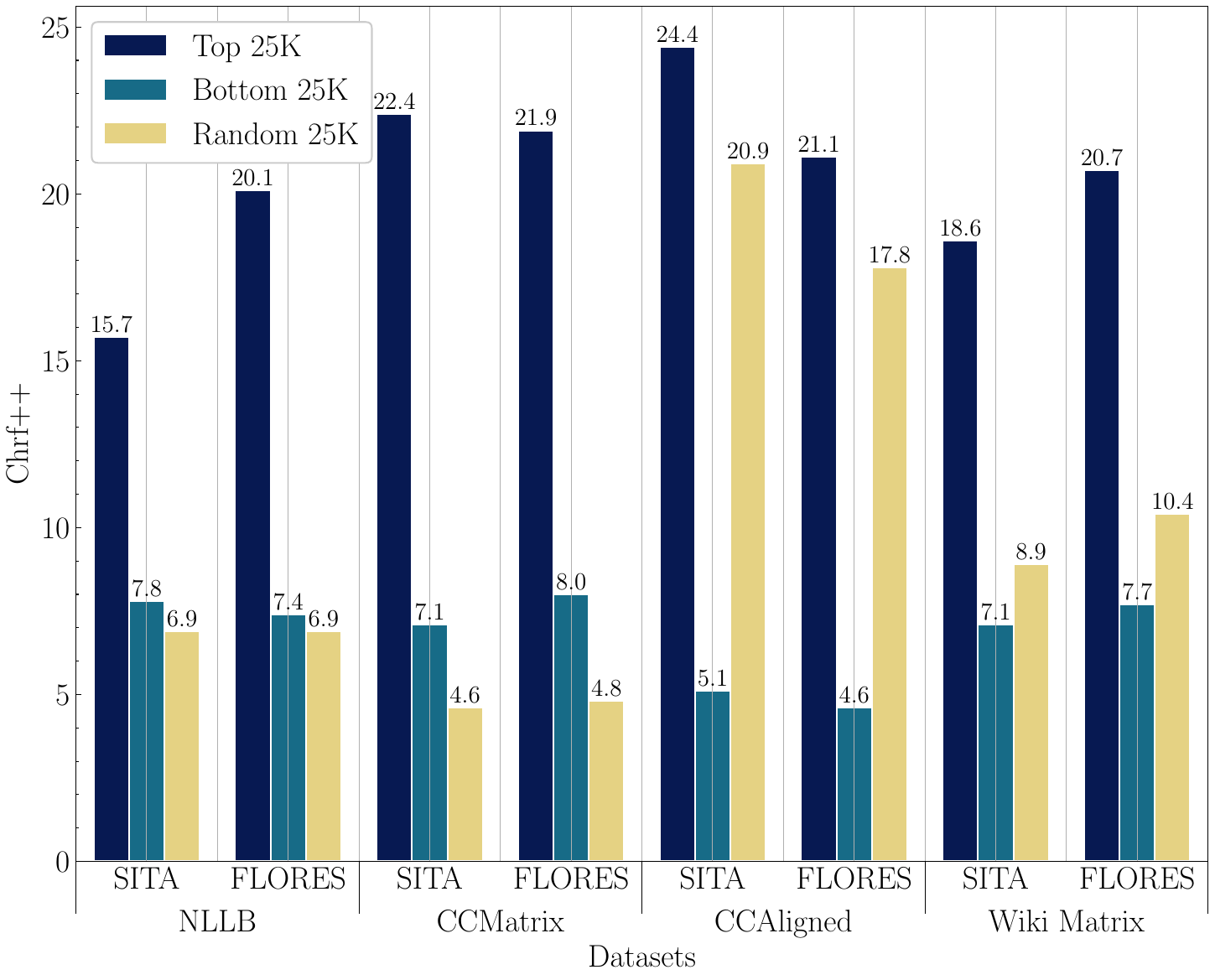}
    \caption{Vanilla-transformer performance trained on Top, Bottom and Random 25K splits of NLLB, CCMatrix, CCAligned and WikiMatrix for \EnSi{} (higher the better).
}
    \label{fig:all_datasets_comp_flores_gov}
\end{figure}

\begin{figure}[tb]
    \centering
    
    \includegraphics[width=0.4\textwidth]{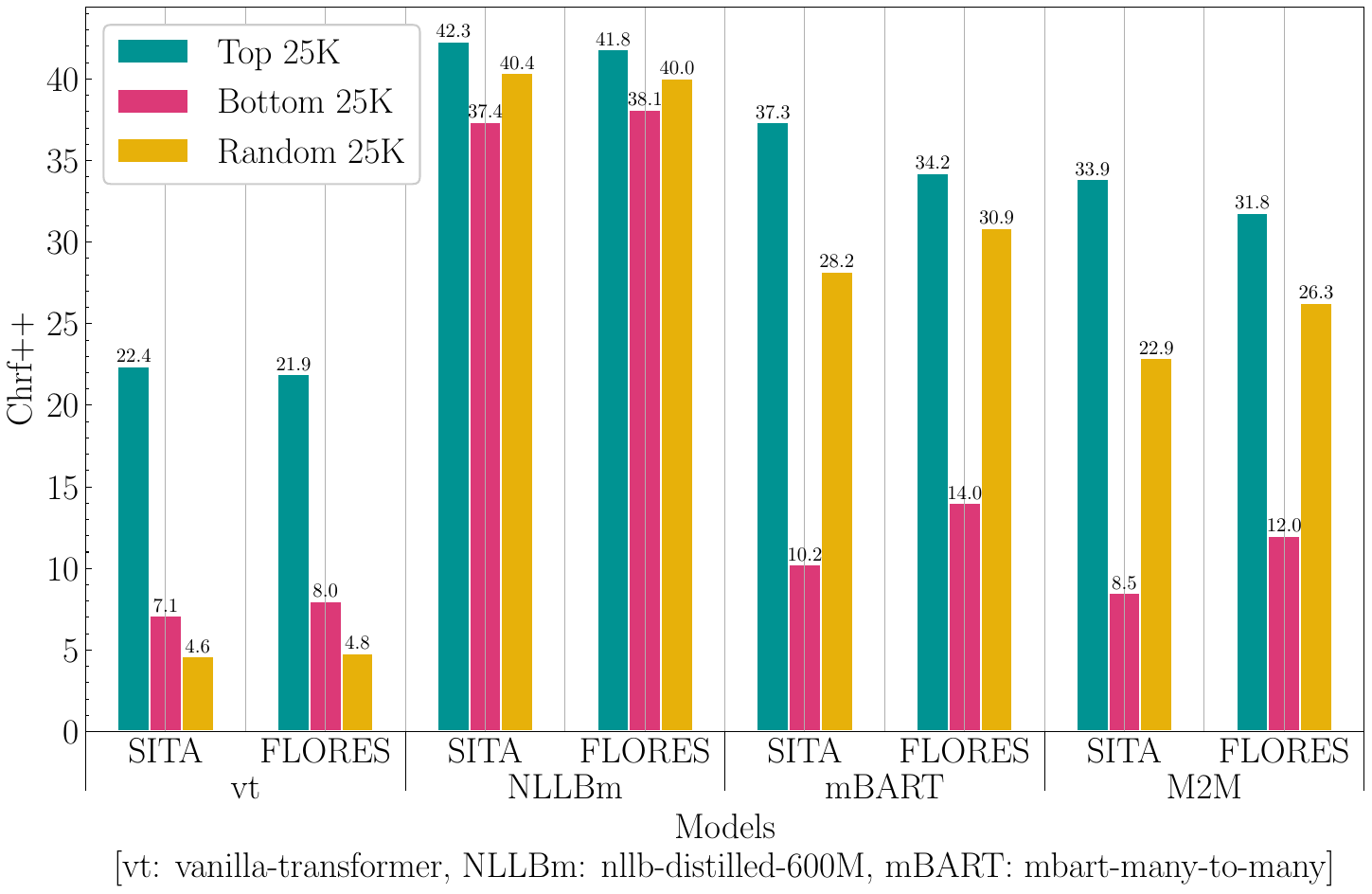}
    \caption{NMT results of different models trained on CCMatrix \EnSi{} top, bottom and average 25K splits.}
    \label{fig:models_comp_flores_gov}
\end{figure}

Further, we observed some qualitative issues in the top 25k splits that are idiocentric to each dataset (or at least more prevalent in a particular dataset than others). 
{CCMatrix} has many untranslated/partially translated pairs.  {WikiMatrix}, on the other hand, has many partial sentences. {CCAligned} has many concatenated lists coming from product advertisements (e.g., cameras, dongles, cables).  Further, this dataset also has a comparatively higher amount of short entries. 
In general, {NLLB} was free of the above faults. However, as touched on in Table~\ref{tab:BadParallel}, the top pairs of NLLB are predominantly religious text with many misalignments. Information in some of the aligned NLLB sentences was not balanced (i.e. one side has more information). 


\begin{figure}[tb]
    \centering
    \captionsetup{justification=centering}
    \includegraphics[width=0.4\textwidth]{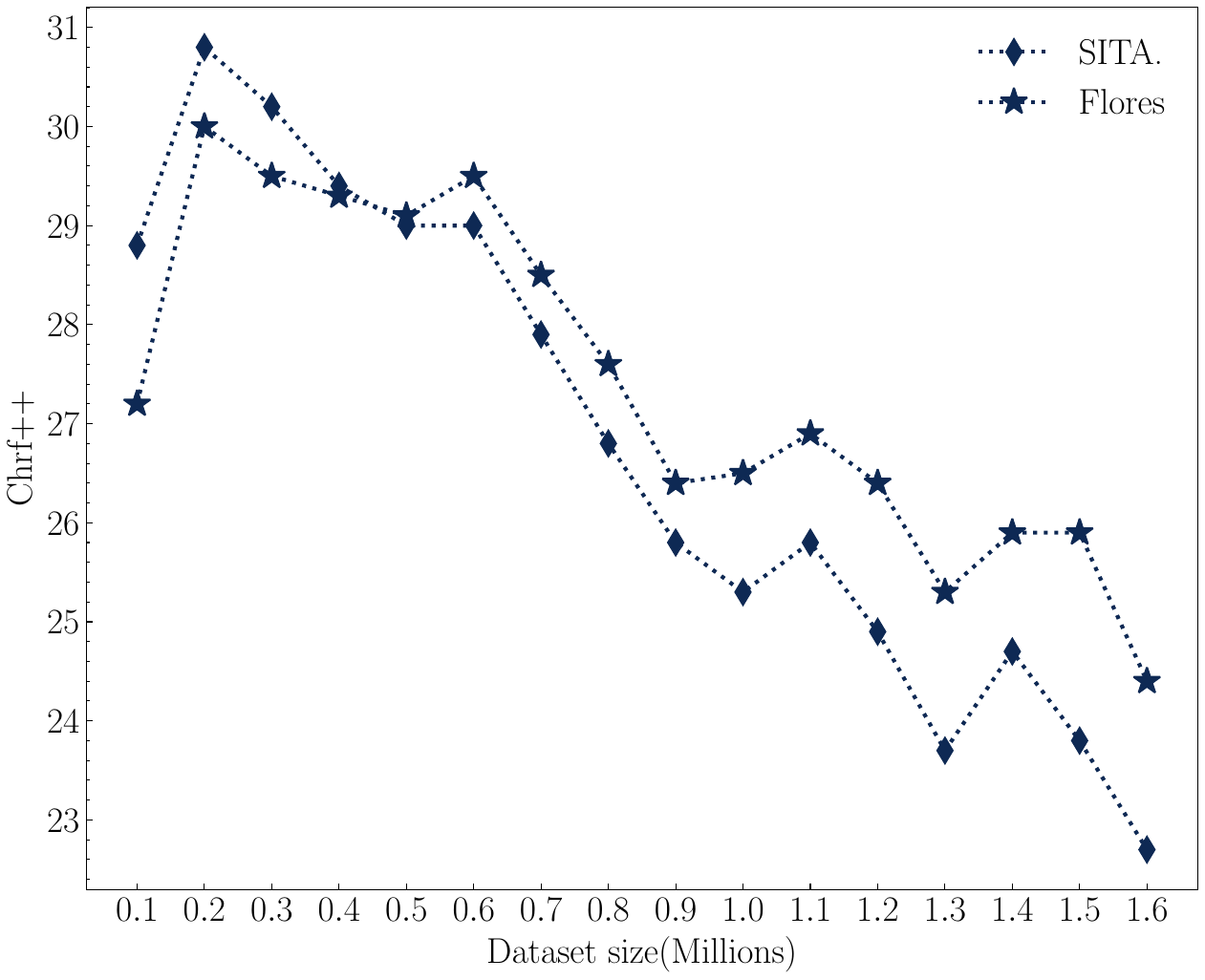}
    \caption{ \RaggedRight NMT results of vanilla transformer model trained on CCMatrix \EnSi{} in jumps of 100K.}
    \label{fig:models_comp_flores}
\end{figure}

The fact that NLLB has more religious text is worth noting because (1) NLLB is presented as a general domain dataset and not one in the religious domain  (2) The phrasing and language used in these religious texts are more archaic than modern. Thus a model trained on the top 25k of NLLB might have a domain bias toward religious text and be unable to handle contemporary language.   

\section{Impact of Corpus Quality on NMT Model Performance}
\label{sec:NMT_model_performance}

In Section~\ref{sec:human_results}, it was evident that different splits of a large web-mined corpus have different levels of quality. To determine whether this quality difference has any impact when it is used to train NMT models, we ran a series of experiments. 

\paragraph{Dataset:}For each corpus, we trained separate NMT models from the top, bottom, and random 25k portions of each of the web-mined \EnSi{} corpora.
We used two separate datasets for testing: \textsc{Flores}-101~\cite{goyal-etal-2022-flores}, and the test set of the SITA parallel corpus~\cite{fernando2020data}. \textsc{Flores} was created from Wikipedia articles, and SITA from government documents of Sri Lanka. 

\paragraph{Baseline Models:}
For \SiTa, \EnSi, and \EnTa,~\citet{thillainathan2021fine, lee-etal-2022-pre} showed that NMT models built on mBART~\cite{tang2020multilingual} outperformed those built on vanilla Transformer models. NMT-specific models such as M2M~\cite{fan2021beyond} and NLLB~\cite{nllb2022} (henceforth referred to as NLLBm, to distinguish from the NLLB dataset) have been shown to be generally better for low-resource languages~\cite{zhu2023multilingual}. However, these models have not been tested for the considered languages. Despite their performance, there is a possibility that the datasets considered in our experiments have already been included in these models~\cite{jacovi-etal-2023-stop}. Thus, we trained vanilla Transformer NMT models with all data splits, and ran an ablation study with CCMatrix \EnSi~for NMT models trained on mBART, NLLB and M2M\footnote{mT5~\cite{xue-etal-2021-mt5} was not used as~\citet{nayak2023leveraging} showed that it lags behind mBART.}. Model and training details are in Appendix~\ref{app:model_details}.

Results were recorded in chrF~\cite{popovic-2015-chrf}, chrF++~\cite{popovic-2017-chrf}, BLEU~\cite{papineni-etal-2002-bleu} and spBLEU~\cite{goyal-etal-2022-flores}.  chrF++ results are used in our discussion. All results are in Appendix~\ref{app:nmt_results}.

Results in Figure~\ref{fig:all_datasets_comp_flores_gov} (Raw result in Table~\ref{tab:fig1_raw_data} in Appendix~\ref{app:nmt_results}) confirm the observations we derived from human evaluation - the top 25k split is significantly better than the other splits. With respect to the SITA test set, the performance ordering of the corpora also tallies with human evaluation results for the Correct (\verb|C|) category:  CCAligned is the best, followed by CCMatrix, WikiMatrix, and NLLB. For \textsc{FLORES} test, CCMatrix is the best, followed by CCAligned, WikiMatrix, and NLLB. Interestingly, despite being created from Wikipedia, WikiMatrix could not beat CCAligned or CCMatrix for \textsc{FLORES}, which was also created from Wikipedia. The lowest result from NLLB could be due to its quality issues, as well as its religious content (see Section~\ref{sec:qualitative_analysis}). Except in CCAligned, both bottom and random splits show roughly similar performance. The high result for the random split in CCAligned correlates with the higher value reported for the \verb|C| category during human evaluation.  

Figure~\ref{fig:models_comp_flores_gov} (Raw result in Table~\ref{tab:fig2_raw_data} in Appendix~\ref{app:nmt_results}) shows how NMT systems built with different pre-trained models perform on CCMatrix \EnSi~data splits. Overall, NMT models built on top of NLLBm show the best performance, followed by mBART and M2M-based models. 
Despite model-wise differences, these results reaffirm that NMT models trained with different splits of the same corpus have different levels of performance. This difference is least pronounced in the NLLBm model. Even in mBART and M2M models, the results gap between top and random splits is minimal, compared to the vanilla transformer model. This confirms that NMT systems built on pre-trained models are more robust to noise in parallel corpora.

\begin{table*}[!htb]
\small
\resizebox{\textwidth}{!}{%
\begin{tabularx}{\textwidth}{|X|l|X|}
\hline
Sentence pair status & Decision & Subsequent action \\
\hline
\hline
Perfect translation (\verb|CC|)  & \verb|ACCEPT| & Keep as it is \\
\hline
\makecell[l]{Acceptable translation, but \En{} and/or \Si{} has\\to be updated (\verb|CN|, \verb|CS| and \verb|CB| in taxonomy)} & \verb|UPDATE| & Update \En{} and/or \Si{}\\
\hline
\makecell[l]{\En~ AND \Si~ both are either meaningless (i.e \verb|NL| or 
\\\verb|WL|), contain repetitive words (eg: No no no), or \\contain very short phrases (\verb|CS|) (e.g. name of a place \\or a person)} & \verb|DELETE| & Keep as it is \\
\hline
\En~ AND \Si~ are meaningful sentences but not related (\verb|X|) & \verb|REWRITE| & \makecell[l]{Add two separate entries - \En{} should be translated\\to \Si, and \Si{} should be translated to \En}\\
\hline
Only \En{} OR \Si{} are meaningful (i.e. one is \verb|NL|, \verb|WL|, \verb|UN|, \verb|CS|) & \verb|REWRITE| & \makecell[l]{Rewrite the un-meaningful side to be the translation\\of the meaningful side}\\
\hline
\end{tabularx}
}
\caption{Decision set employed for manual cleaning of the corpus ({We remove ones marked as \texttt{DELETE} from the corpus before using it for NMT training.})}
\label{tab:guidelines}
\end{table*}

\begin{table}[!ht]
\resizebox{\linewidth}{!}{%
\begin{tabular}{lrrrrr}
\hline
\multirow{2}{*}{\textbf{Decision}} &\multicolumn{2}{c}{\textbf{\EnSi}} &\multicolumn{2}{c}{\textbf{\EnTa}} \\
\textbf{}&\textbf{Total Sentences} &\textbf{\%} &\textbf{Total Sentences} &\textbf{\%} \\
\hline
Accept &4813 &17.13 &6621 &24.70 \\
Update &14852 &52.87 &15047 &56.14 \\
Re-write &8148 &29.01 &4858 &18.13 \\
Delete &277 &0.99 &275 &1.03 \\
\hline
\textbf{Total} &\textbf{28090} &\textbf{} 
&\textbf{26801} &\textbf{} \\
\hline
\end{tabular}}
\caption{Summary of translator decisions}\label{tab:nllb_cleaning_results}
\end{table}

These findings naturally lead to the question `\textit{what would happen to the NMT performance if the dataset size is gradually increased beyond 25k?}'. To answer this question, we trained vanilla transformer-based NMT models, by gradually increasing the size of the CCMatrix \EnSi{} corpus up to 1.6M\footnote{We did not go above 1.6M due to resource limitations.}. Figure~\ref{fig:models_comp_flores} shows the results (Raw results are in Table~\ref{tab:fig3_raw_data} in Appendix~\ref{app:nmt_results}). Despite fluctuations, when the training dataset size increases, the results gradually decrease. Also note that for this corpus, the peak result is achieved when the training set is 200k. This number may vary from corpus to corpus\footnote{Thus, although we used 25k as our portion size, this number should not be taken as a universal cut-off value.}.

We also trained vanilla Transformer models from the full CCMatrix, CCAligned and WikiMatrix for \EnSi{}. Corresponding chrF++ results are 17.8, 41.7 and 17.3 (respectively) for SITA and 20.4, 31.7 and 19 (respectively) for \textsc{FLORES}. Comparing these values with those in Figure~\ref{fig:models_comp_flores_gov} shows that for some corpora, training an NMT model just with the top 25k split is better than using the full corpus.

\section{Impact of Corpus Cleaning}
\label{sec:impact_manual_cleaning}
\subsection{Process and Human Evaluation}
Creating high-quality corpora is a challenging task, especially for low-resource languages. In this context, employing human translators to clean web-mined corpora can be considered an alternative to creating parallel corpora from scratch. 


In order to determine the benefit of corpus cleaning, we cleaned the top 25k of NLLB \EnSi{} and \EnTa{} corpora. 11~\EnSi{} translators and 16~\EnTa{}  translators were used for this task\footnote{In the case of the two translators who were involved in the corpus evaluation in addition to cleaning, we made sure not to (re)assign the samples they evaluated.}. Details of translator selection and training are shown in Table~\ref{tab:cc_translators} of Appendix~\ref{app:appendix_human_details}. The translators were asked to first indicate the decision they took on a given sentence pair. The set of decisions and subsequent actions expected by the translator are given in Table~\ref{tab:guidelines}. 

Due to rewrites and deletes, the resulting corpus now has a final cleaned sentence pair count of 27,813 for \EnSi{} and 26,526 for \EnTa. Table~\ref{tab:nllb_cleaning_results}  shows the statistics of decisions taken by the translators. We see a significant number of updates, which confirms that the original corpus had more pairs falling into the \verb|C| category. The lesser, but significant rewrites and very low count of deletes confirm that the \verb|E| category was relatively small.

Recall that we conducted a human evaluation of 250 random samples from this portion of the NLLB corpus for both \EnSi{} and \EnTa{} corpora. Each of these 250 samples was cleaned by three separate translators, while each sentence in the rest of the corpus was cleaned by a single translator.  
The 250 sentences of the top 25k portion of NLLB \SiTa{} corpus that were used for quality estimation were also cleaned by three translators. The last three rows in Table~\ref{tab:human_eval_results} show this result. We see a significant drop in error (\verb|E|) categories and a significant increase in \verb|CC| category. However, human data cleaning has not produced a perfect result - had it been perfect, we should have seen 100\% for \verb|CC+CS| categories.

We manually reviewed the cleaned \EnSi{} translation pairs that did not fall into \verb|CC| or \verb|CS| categories to identify why they were not cleaned to be perfect translation pairs. Our observations are as follows:

\begin{itemize}
    \item  NLLB has a high concentration of religious text. Jargon and structure used in the religious text are very different to contemporary vernacular. Some translators found it difficult to find equivalent wording in Sinhala for the religious-specific language. 
    \item Some English sentences had structural issues. Some translators have not bothered to fix these structural issues and have simply translated that ill-formed English sentence into Sinhala.
    \item  In the cases where the English sentence is partial (e.g. interrupted utterance), translating it to Sinhala was difficult due to differences in grammatical word ordering.
    \item  Some English sentences that discuss ideas that are rooted in Western culture had no concise way of translating (e.g. I am taking her on a date).
    \item Spelling errors\footnote{Sinhala does not have a reliable spell corrector, and many errors can be easily overlooked~\cite{sonnadara2021sinhala}.}, errors caused due to overlooking punctuation errors. 

\end{itemize}

Table~\ref{tab:cc_translator_durations} in Appendix~\ref{app:corpus_cleaning} shows the time taken by translators for the corpus cleaning task. To produce 28,090 sentences from the noisy 25k~\EnSi{} corpus, the translators have collectively spent a total of 853:18(hr:m). On average, this is 1.8 minutes per sentence pair. To prepare a sample of a hundred sentence pairs, an average duration of 3hrs 3 minutes with a standard deviation of 1hr and 9 minutes was taken. Cleaning of 25k~\EnTa{} sample produced 26,801 sentences consuming a total duration of 539:52 (hr:m). On average per sentence, the duration spent was 1.2 minutes. The average duration spent for a hundred sentences was 3hrs 47minutes with a standard deviation of 3hrs 57minutes. In both instances, the standard deviation is noticeably high. This is due to the individual capabilities/circumstances of translators or even a translator wrongly recording time(see  Appendix~\ref{app:corpus_cleaning}), it could also be due to the quality of the dataset portion received by a translator and the translator's judgment on the action to be carried out on a given sentence pair.  This assumption is strengthened by results in Table~\ref{tab:cc_translatorwise_per} in Appendix~\ref{app:corpus_cleaning}  - there is a high variance in the actions selected by the translators.

To see if cleaning a web-mined corpus is more effective than translating a source from scratch, we selected three translators from the corpus cleaning task, gave them 100 sentences from NLLB \EnSi{} corpus (that they had not seen before), and asked them to translate from scratch. We compared the time they took for the fresh translation and corpus cleaning. 

As per Table~\ref{tab:cc_vs_translation_durations} in Appendix~\ref{app:corpus_cleaning}, corpus cleaning on average took 14 minutes less than fresh translation. However, the time taken to clean a corpus may vary depending on its quality. For the sake of completion, the freshly translated 100 sentences were evaluated by evaluators. \verb|CC|, \verb|CN|, and \verb|CB| percentages are 57.00\%, 10.67\% and 32.33\% respectively, which are on par with corpus cleaning results.

\subsection{Impact on NMT Performance}
\begin{figure}[tb]
    \centering
        \includegraphics[width=0.35\textwidth]{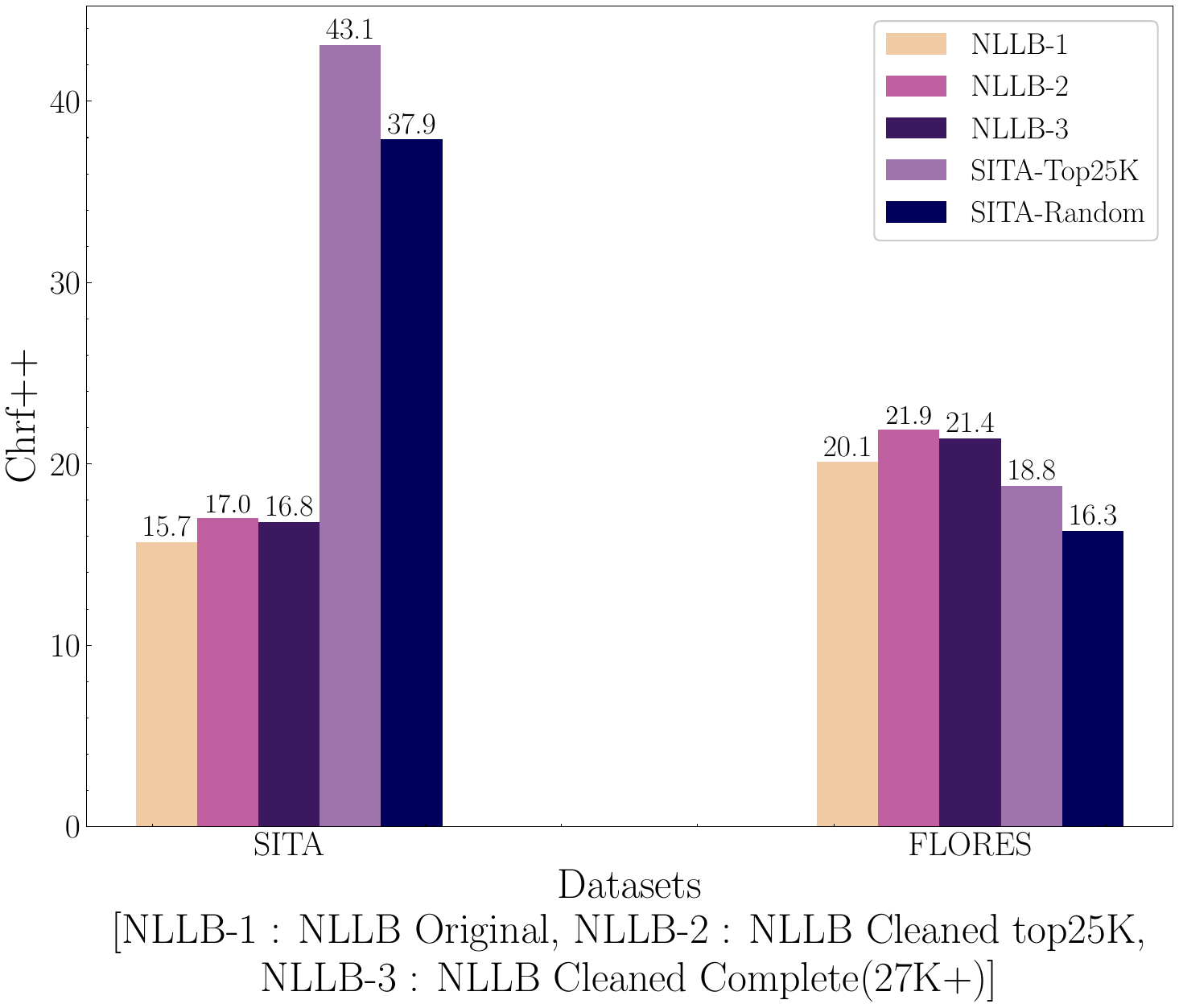}
    \caption{Vanilla transformer results for \EnSi{} original NLLB Top 25K, NLLB cleaned Top 25K, NLLB cleaned full(27K+), SITA Top 25K, and SITA Random 25K.}
    \label{fig:gov_v_nllb_cleaned}
\end{figure}

\begin{figure}[tb]
    \centering
        \includegraphics[width=0.35\textwidth]{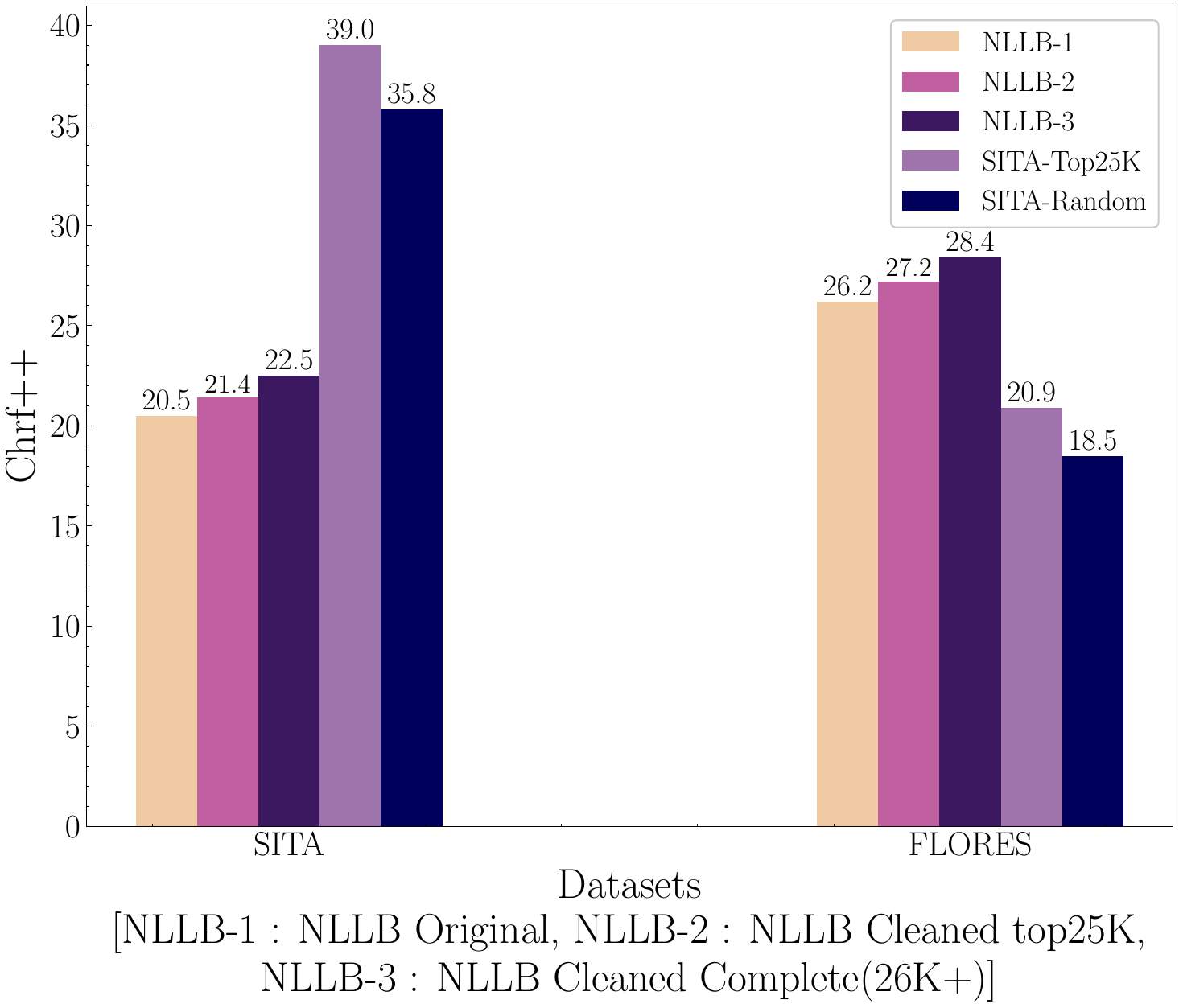}
    \caption{Vanilla transformer results for \EnTa{} original NLLB Top 25K, EnTa NLLB cleaned Top 25K, EnTa NLLB cleaned full(26K+), EnTa SITA Top 25K, and EnTa SITA Random 25K. }
    \label{fig:gov_v_nllb_enta}
\end{figure}

For both \EnSi{} and \EnTa{}, we built NMT models from the fully cleaned NLLB corpus, its top 25k, as well as from the random and the top 25k splits of SITA corpus.
Figures~\ref{fig:gov_v_nllb_cleaned} and~\ref{fig:gov_v_nllb_enta} show the respective results. Raw results tables are in Table~\ref{tab:fig4_raw_data} in Appendix~\ref{app:nmt_results}\footnote{Though we report result of SITA training set on SITA test set, this is misleading due to both coming from same corpus.}. For both language pairs, cleaned NLLB top 25k corpus beats the uncleaned version for SITA and \textsc{FLORES} test sets. But, compared to human effort to clean the corpus, this gain cannot be justified.

As per Figures~\ref{fig:gov_v_nllb_cleaned} and~\ref{fig:gov_v_nllb_enta}, the top 25k split of both web-mined corpora performed better than SITA top 25k split for \textsc{FLORES} test set. \citet{nayak2023leveraging} showed that NMT results could be affected by domain divergence. To determine whether this drop in SITA result is due to domain divergence, we calculated JS Divergence between different corpora (see Table~\ref{tab:domain_divergence} in Appendix~\ref{app:dom_Div}). The divergence between SITA and \textsc{FLORES} is the highest, but it is only slightly higher (0.1 points) than that of CCAligned. However, CCAligned result for \textsc{FLORES} is 2.3 chrF++ points higher than SITA. Therefore it is safe to assume the low performance of SITA may not be due to domain divergence, but due to its quality. However, the high domain divergence between NLLB and SITA is noteworthy. We remind the reader that we noticed NLLB having higher amounts of religious content (see Section~\ref{sec:qualitative_analysis}).

The full NLLB cleaned \EnSi{} corpus of 27k+ lags behind the top 25k split. Similarly, for both language pairs, SITA random 25k lags behind the top 25k. 

\section{Impact of Embedding Technique}

\begin{figure}[!htb]
    \centering
        \includegraphics[width=0.35\textwidth]{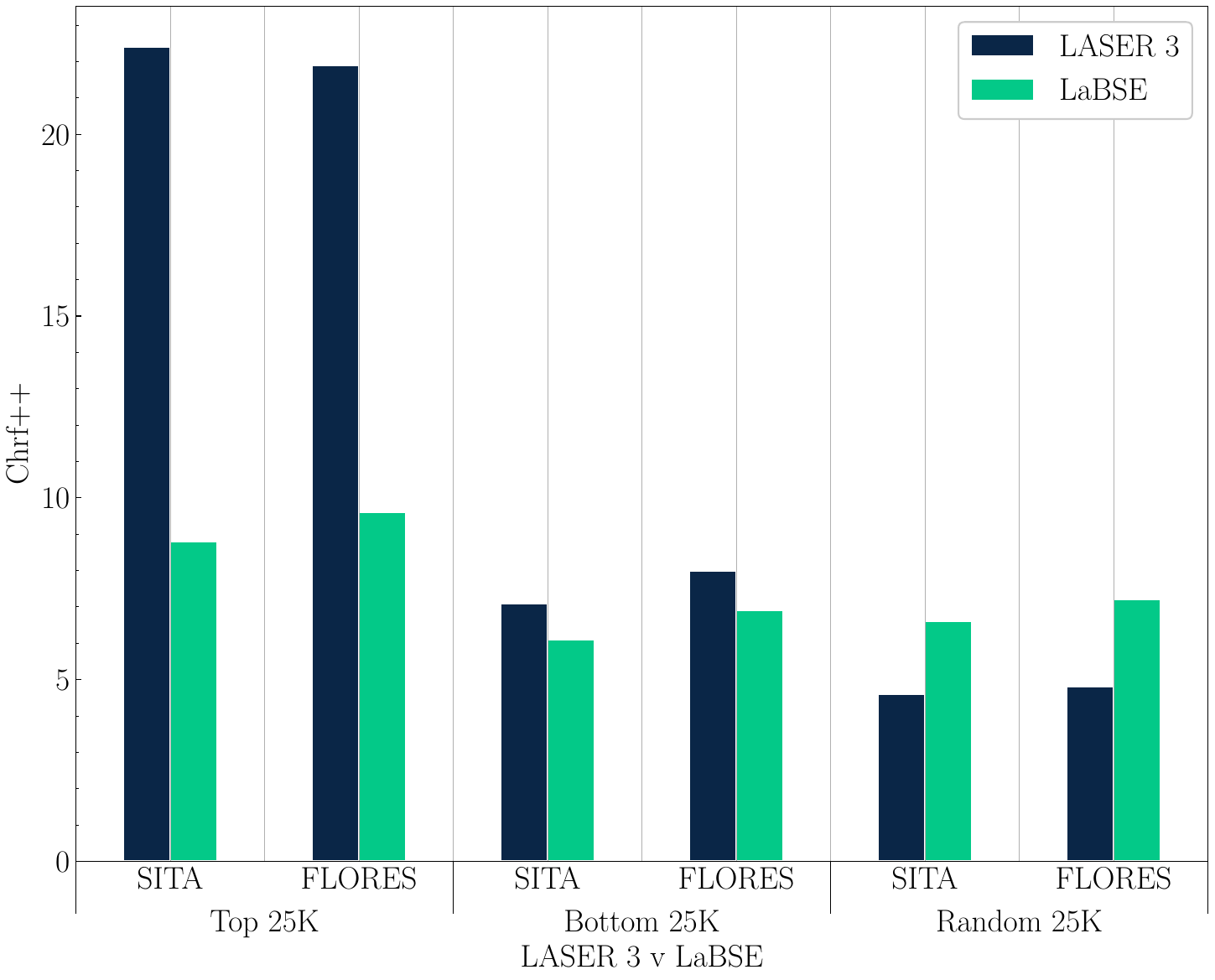}
    \caption{NMT Results on CCMatrix \EnSi{} Top, Bottom and Random 25K for LASER-3 or LaBSE.}
    \label{fig:labse_v_laser3}
\end{figure}

We used LASER-3 for ranking sentence pairs. The other commonly used measurements are LaBSE and XLM-R~\citep{conneau-etal-2020-unsupervised}. mBERT is another option, however, Sinhala is not included in this model. In fact,~\citet{fernando2023exploiting} compared LASER-1, LaBSE, and XLM-R embedding performance for sentence alignment and reported that LaBSE is superior to the other two. To determine whether the embedding technique has a noticeable impact, we ranked the CCMatrix \EnSi{} corpus using LaBSE and XLM-R. Then we selected the top, bottom, and random splits from this corpus and trained vanilla transformer-based NMT models. Figure~\ref{fig:labse_v_laser3} shows the comparison (Full result in Table~\ref{tab:fig5_raw_data} of Appendix~\ref{app:nmt_results}). XLM-R result is close to zero, so is not shown. Overall, the top 25k ranked by LASER-3 has a higher result than the other two. However, note that for a given language pair, the actual result may depend on the language representation in the model and the characteristics of the corpus. Thus, for a new pair of languages, it is worthwhile to experiment with different embedding models.  


\section{Conclusion}
We presented a fine-grained evaluation of the quality of web-mined corpora for three low-resource language pairs. We showed that the quality of such corpora significantly varies across different portions. Our findings also indicate that simply using the highest quality portion of a web-mined corpus yields NMT results that may be on par with human-curated corpora in some instances. However, we are wary of further cleaning this top portion in hopes of better results, as the result gains do not justify the required human effort. Project artefacts are released and the details are shared in the project GitHub\footURL{https://github.com/nlpcuom/quality-matters}.

For our analysis, we considered the web-mined corpora without any pre-processing. If they were pre-processed (say) to remove duplicates, short phrases, or text in the wrong language, the performance of the embedding techniques may vary. We plan to investigate this in future. We also plan to expand this analysis to other low-resource languages.

\section*{Limitations}
Our evaluation involves only three languages. This was inevitable because these are the only languages we had provisions to find human translators to carry out a meaningful evaluation.
Due to financial constraints, we could carry out data cleaning only for the \EnSi{} and \EnTa{} portions of the NLLB corpus. For the NLLB cleaning task, we reviewed only the first 100 sentences produced by the human translators. Therefore this corpus could still have some noise. From each corpus, we reviewed only 750 sentences. While this number is much larger than what~\citet{kreutzer-etal-2022-quality} considered, it may still not be representative enough. Due to computing resource constraints, we could not train NMT models with all pre-trained models or train NMT models for various sizes of all parallel corpora. Our technique works only for languages included in embedding models such as LASER, LaBSE, and XLM-R.

\section*{Ethics Statement}
We used publicly available parallel corpora that are free to use. \citet{fernando2020data} provided their dataset. We paid all the translators according to the government's stipulated rates. Before assigning them to the task, they were given a pilot to try out. They were given the chance to decide whether they were adequately compensated for their efforts. We only collected personal information that is needed for us to determine their suitability for the task and to arrange their payment. None of these personal details has been publicly released. More details are in the Appendix~\ref{app:corpus_cleaning}. As mentioned under limitations, we could not manually review the corpus cleaned by translators. While they fixed the issues in a publicly available corpus, we cannot guarantee that the cleaned corpus does not have any unnecessary content that was not there in the original corpus.  


\section*{Acknowledgements}
This work was funded by the Google Award for Inclusion Research (AIR) 2022 received by Surangika Ranathunga and Nisansa de Silva.

\bibliography{anthology,custom}
\bibliographystyle{acl_natbib}

\clearpage
\newpage

\appendix

\section{Parallel Corpora used in the Study}
\label{app:corpora}

All following artefacts were used consistent with their intended use when and where it was specified. The creators of the respective artefacts have checked whether their data contains any information that uniquely identifies individual people or offensive content. In the cases where the data is \textit{updated} or \textit{re-written} by translators as discussed in Section~\ref{sec:NMT_model_performance}, the guideline discussed in Appendix~\ref{app:corpus_cleaning} ensured that no information that uniquely identifies individual people or offensive content is inserted. The licences and terms of usage of the artefacts are as discussed in each of the cited sources below.

\paragraph{CCAligned~\citep{el-kishky-etal-2020-ccaligned}}is a dataset created using 68 snapshots of CommonCrawl\footURL{http://commoncrawl.org/}. Document alignment was done using FastText LangID~\citep{joulin2016fasttext,joulin-etal-2017-bag} by mapping documents with the same URL but different language codes. The alignments were then refined using  LASER embeddings~\citep{artetxe-schwenk-2019-massively}.

\paragraph{WikiMatrix~\citep{schwenk-etal-2021-wikimatrix}}is a parallel corpus mined from Wikipedia. It has 135M parallel sentences in 1620 language pairs (85 languages). 34M of these are aligned with English. Duplicates have been removed after sentence splitting. FastText LangID has been used to identify the languages of the text and then LASER has been used to identify bitext.

\paragraph{CCMatrix~\citep{schwenk-etal-2021-ccmatrix}}was created using snapshots of CommonCrawl. It contains around 4.5 billion parallel sentences across 576 language pairs. In building CCMatrix, it was assumed the aligned sentence could appear anywhere on CommonCrawl. Thus, margin-based mining~\citep{artetxe-schwenk-2019-margin} was used for sentence alignment.

\paragraph{NLLB~\citep{nllb2022}}was released along with a translation model of the same name. This dataset contains: (1) public primary bitext collected from various sources, (2) bitext mined with LASER-3 teacher-student training~\citep{heffernan-etal-2022-bitext}, and (3) Backtranslated bitext created from the monolingual corpus.


\section{Human Evaluator Details}
\label{app:appendix_human_details}

Table~\ref{tab:corp_evaluators} provides details about the human participants involved with the evaluation task.
{They all possess a minimum of one year of prior experience in translation.}  They are from Sri Lanka. We advertised for this work via social media. This set of translators was selected from a larger pool via a small test, by giving them ten sentences to translate.  

\begin{table}[!ht]
\resizebox{\linewidth}{!}{%
\begin{tabular}{lcl}
\hline
\textbf{Name} &\textbf{Experience} &\textbf{Qualification} \\
&\textbf{(Years)}& \\
\hline
\multicolumn{3}{l}{\textbf{En - Si}} \\
\hline
Translator1 &1 &BA in German Language \\
Translator1 &2 &BA (Hons) Sinhala Sp. \\
Translator2 &2 &BA (Hons) in Translation Studies \\
Translator3 &1 &BA (Hons) in Translation Studies \\
Translator4 &1 &BA (Hons) in Translation Studies \\
Translator5 &4 &BA (Hons) in Translation Studies \\
Translator6 &2 &BA (Hons) in Translation Studies \\
\hline
\multicolumn{3}{l}{\textbf{En - Ta}} \\
\hline
Translator7 &7 &BSc in Agriculture \\
Translator8 &12 &B.Sc. Applied Mathematics and Computing \\
&&PGD in Professional Practice\\
&&in English \\
Translator9 &5 &BSc (Hons) Engineering \\
Translator10 &3 &MBBS \\
Translator11 &5 &BA \\
\hline
\multicolumn{3}{l}{\textbf{Si - Ta}} \\
\hline
Translator13 &2 &BA (Hons) in Translation Studies \\
Translator14 &2 &BA (Hons) in Translation Studies \\
Translator15 &20 &Diploma in Translation and Interpretation \\
\hline
\end{tabular}
}
\caption{Details of  Translators involved in Corpus Evaluation task}\label{tab:corp_evaluators}
\end{table}

\begin{figure*}[!htb]
    \centering        \includegraphics[width=\textwidth]{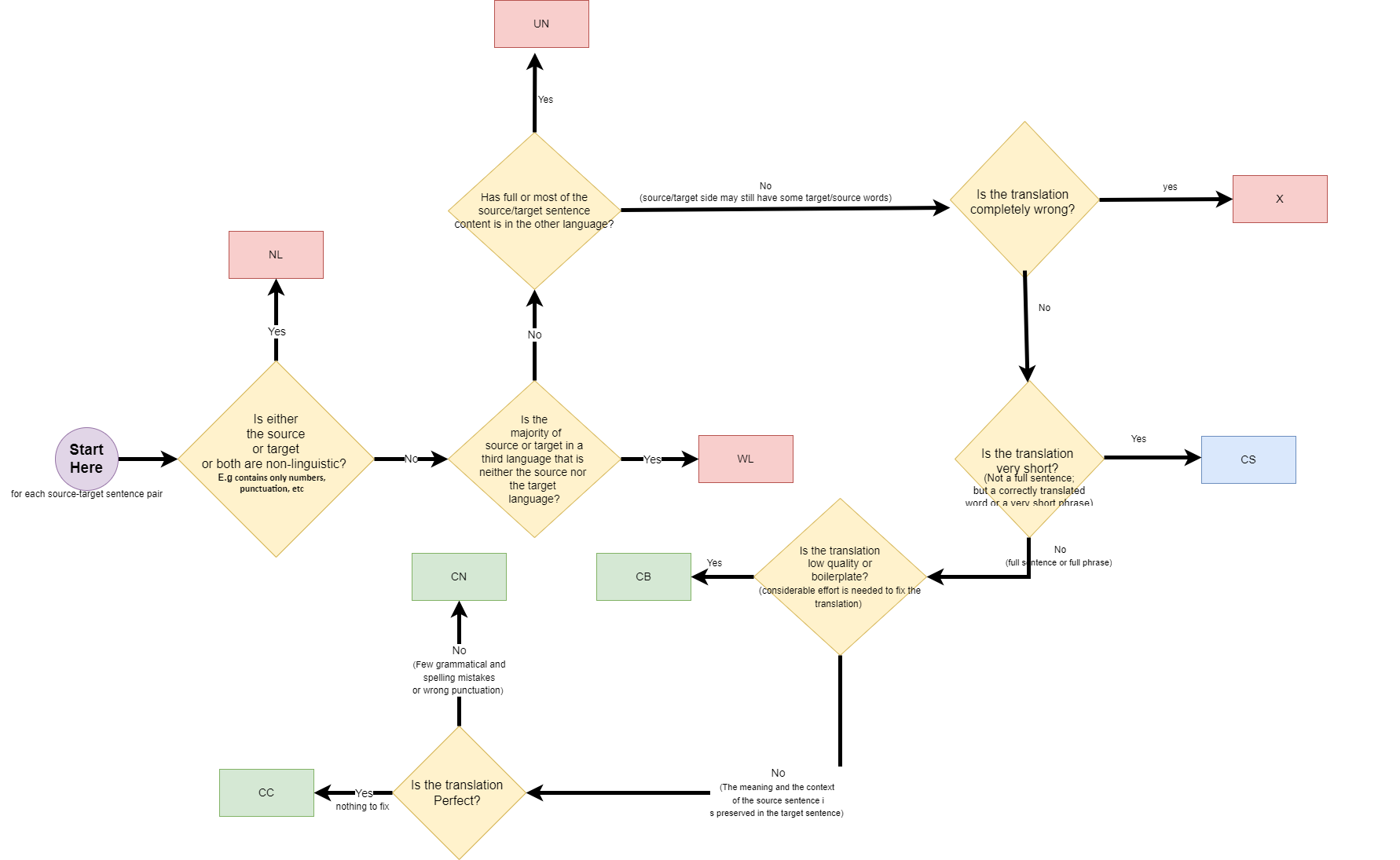}
    \caption{Flow-chart for Corpus Evaluation}
    \label{fig:flow_chart}
\end{figure*}

A flow-chart (See Figure~\ref{fig:flow_chart}) was prepared to explain the evaluation task. Then they were given a pilot set to practice the task. We evaluated their work, refined the guidelines and provided them with the final instructions along with a demonstration video. They were paid for each sentence they evaluated. Before assigning work, we informed them of the rates. Thus, based on the time taken for the pilot task, the translators were given the option to decide whether they wanted to continue with the full task under the proposed payment rates.
Table~\ref{tab:correlation} contains the raw data used for the Pearson correlation study.

\begin{table*}[!htb]
    \centering
\resizebox{\textwidth}{!}{%
\tiny
    \begin{tabular}{lll|d{2.2}d{2.7}|d{2.2}d{2.7}|d{2.2}d{2.7}}
        \hline
        \multicolumn{3}{l|}{\multirow{2}{*}{Dataset}} & \multicolumn{2}{c|}{Eval1 v Eval2}& \multicolumn{2}{c|}{Eval2 v Eval3} & \multicolumn{2}{c}{Eval1 v Eval3}\\
        & & & \CorrelationMetric & \CorrelationMetric & \CorrelationMetricEnd \\
       \hline
        \multirow{6}{*}{CC Align}  
        & \multirow{3}{*}{\EnSi}  &Top & 0.31 & 4.22E-07 & -0.28 & 5.79E-06 & -0.04 & 0.51 \\   
        & &Random & 0.19 & 0.00 & 0.19 & 0.00 & 0.28 & 8.85E-06 \\
        & &Bottom & 0.78 & 6.31E-52 & 0.74 & 2.61E-45 & 0.68 & 9.42E-36 \\ 
        \hhline{~--------}
        & \multirow{3}{*}{\EnTa}  &Top & 0.08 & 0.23 & 0.00 & 0.96 & 0.54 & 1.41E-20 \\      
        & &Random & 0.55 & 7.88E-21 & 0.49 & 7.61E-17 & 0.58 & 6.33E-24 \\
        
        & &Bottom & 0.60 & 1.99E-25 & 0.72 & 1.37E-41 & 0.73 & 2.21E-42 \\ 
        \hline

         \multirow{6}{*}{Wikimatrix}  
        & \multirow{3}{*}{\EnSi}  &Top & 0.19 & 0.00 & 0.27 & 1.13E-05 & 0.36 & 6.36E-09 \\   
        & &Random & 0.34 & 3.04E-08 & 0.38 & 7.43E-10 & 0.29 & 2.14E-06 \\
        & &Bottom & 0.37 & 1.06E-09 & 0.52 & 4.44E-19 & 0.50 & 2.22E-17 \\ 
        \hhline{~--------}
        & \multirow{3}{*}{\EnTa}  &Top & 0.65 & 4.49E-31 & 0.76 & 1.10E-47 & 0.68 & 9.28E-36 \\      
        & &Random & 0.35 & 1.42E-08 & 0.17 & 0.01 & 0.37 & 2.38E-09 \\
        
        & &Bottom & 0.53 & 1.34E-19 & 0.56 & 2.63E-22 & 0.67 & 9.30E-34 \\ 
        \hline 
        
        \multirow{9}{*}{CC Matrix}  
        & \multirow{3}{*}{\EnSi}  &Top & 0.35 & 1.26E-08 & 0.52 & 7.35E-19 & 0.44 & 2.28E-13 \\   
        & &Random & 0.43 & 7.19E-13 & 0.55 & 1.91E-21 & 0.37 & 2.05E-09 \\
        & &Bottom & 0.33 & 1.16E-07 & 1.00 & 0.00 & 0.33 & 1.16E-07 \\ 
        \hhline{~--------}
        & \multirow{3}{*}{\EnTa}  &Top & 0.73 & 1.66E-42 & 0.75 & 4.23E-47 & 0.86 & 9.73E-73 \\      
        & &Random & 0.59 & 9.45E-25 & 0.43 & 1.06E-12 & 0.49 & 1.23E-16 \\
        
        & &Bottom & 0.51 & 8.11E-18 & 0.42 & 7.14E-12 & 0.62 & 6.13E-28 \\ 
        \hhline{~--------}
        & \multirow{3}{*}{\SiTa}  &Top & 0.10 & 0.11 & 0.65 & 1.04E-31 & 0.16 & 0.03 \\      
        & &Random & 0.63 & 7.61E-29 & 0.66 & 1.32E-32 & 0.55 & 1.46E-21 \\
        
        & &Bottom & 0.61 & 2.50E-27 & 0.74 & 6.29E-45 & 0.65 & 3.89E-31 \\ 
        \hline 

        \multirow{9}{*}{NLLB}  
        & \multirow{3}{*}{\EnSi}  &Top & 0.57 & 2.80E-23 & 0.51 & 1.11E-17 & 0.63 & 1.39E-29 \\   
        & &Random & 0.38 & 4.44E-10 & 0.32 & 2.80E-07 & 0.47 & 2.88E-15 \\
        & &Bottom & 0.51 & 4.31E-18 & 0.46 & 1.58E-14 & 0.30 & 1.38E-06 \\ 
        \hhline{~--------}
        & \multirow{3}{*}{\EnTa}  &Top & 0.64 & 3.42E-30 & 0.69 & 4.23E-36 & 0.68 & 3.60E-35 \\      
        & &Random & 0.51 & 1.01E-17 & 0.45 & 8.71E-14 & 0.58 & 1.89E-23 \\
        
        & &Bottom & 0.57 & 4.58E-23 & 0.63 & 2.19E-29 & 0.69 & 2.01E-36 \\ 
        \hhline{~--------}
        & \multirow{3}{*}{\SiTa}  &Top & 0.56 & 1.31E-21 & 0.58 & 1.15E-23 & 0.64 & 1.22E-30 \\      
        & &Random & 0.60 & 3.85E-26 & 0.66 & 4.70E-32 & 0.67 & 2.61E-34 \\
        
        & &Bottom & 0.54 & 1.96E-20 & 0.65 & 2.59E-31 & 0.66 & 2.59E-32 \\ 
        \hline

    \end{tabular}
    }
    \caption{Raw results used for Pearson correlation study for agreement between evaluators (Eval) on 250 samples for \EnSi, \EnTa, and \SiTa}    
    \label{tab:correlation}
\end{table*}


\section{Web-mined Corpus Cleaning}
\label{app:corpus_cleaning}
To clean the top 25k sentences from the NLLB corpus, translators were selected following the same procedure described in Section~\ref{sec:quality_est_human}. Table~\ref{tab:cc_translators} gives details about the human participants involved with the NLLB cleaning task.

\begin{table}[!ht]
\resizebox{\linewidth}{!}{%
\begin{tabular}{lcl}
\hline
\textbf{Name} &\textbf{Experience} &\textbf{Qualification} \\
&\textbf{(Years)}& \\
\hline
\multicolumn{3}{l}{\textbf{En - Si}} \\
\hline
Translator16 &4 &BA (Special) in Translation Studies \\
Translator17 &1 &BA (Hons) in Translation Studies \\
Translator18 &1 &BA (Hons) Business and Academic Chinese \\
Translator19 &4 &MBBS \\
Translator20 &1 &BA (Hons) in Translation Studies \\
Translator21 &2 &BA (Hons) in Translation Studies \\
Translator22 &1 &BEng (Hons) Technology (Mech Eng) \\
Translator23 &3 &BA (Special) in Translation Studies \\
Translator24 &1 &BA(Hons) in French Language and Literature \\
Translator25 &2 &BA (Hons.) in Translation Studies \\
Translator26 &3 &Certificate in Effective English \\
\hline
\multicolumn{3}{l}{\textbf{En - Ta}} \\
\hline
Translator27 &6 &BSc (Hons) in Information Technology \\
Translator28 &1 &Bsc (Hons) Business Information System \\
Translator29 &4 &BA (Hons) in Translation Studies \\
Translator30 &1 &BSc Environmental Conservation and Management \\
Translator31 &1 &Bachelor of Unani Medicine and Surgery \\
Translator32 &4 &Bachelor of Information Technology \\
Translator33 &1 &BSc (Hons) Eng. sp. in Computer Science and Eng. \\
Translator34 &1.5 &BSc (Hons) Eng. sp. in Chemical and Process Eng. \\
Translator35 &3 &BSc (Hons) in Town and Country Planning \\
Translator36 &2.5 &B.Tech in Chemical Engineering \\
Translator37 &1 &BSc (Hons) in Nursing \\
Translator38 &4 &BA (Hons.) in Translation Studies \\
Translator39 &8 &Master of Business Management \\
Translator40 &2 &BA (Hons.) in Translation Studies \\
\hline
\multicolumn{3}{l}{\textbf{Si - Ta}} \\
\hline
Translator41 &6 &BA (Hons) in Translation Studies \\
Translator42 &1 &BA in Social Sciences \\
Translator43 &35 &MA - Linguistic \\
\hline
\end{tabular}}
\caption{Details of Translators Involved in Corpus Cleaning Task}\label{tab:cc_translators}
\end{table}
\begin{table*}[ht]  
\centering
\scriptsize
\resizebox{\textwidth}{!}{%
\begin{tabular}{lrrrrrrrrr}
\hline
\multirow{2}{*}{\textbf{Translator}} &\multirow{2}{*}{\textbf{Total Sentence-pairs}} &\multicolumn{2}{c}{\textbf{Accept}} &\multicolumn{2}{c}{\textbf{Update}} &\multicolumn{2}{c}{\textbf{Re-write}} &\multicolumn{2}{c}{\textbf{Delete}} \\
\hhline{~~--------}
& &\makecell[l]{\textbf{Decision}\\\textbf{Count}} & \textbf{\%} &\makecell[l]{\textbf{Decision}\\\textbf{Count}} &\textbf{\%} &\makecell[l]{\textbf{Decision}\\\textbf{Count}}&\textbf{\%} &\makecell[l]{\textbf{Decision}\\\textbf{Count}} &\textbf{\%} \\
\hline
Translator16 &652 &29 &4.45 &540 &82.82 &74 &11.35 &9 &1.38 \\
Translator17 &1523 &414 &27.18 &664 &43.60 &403 &26.46 &42 &2.76 \\
Translator18 &3402 &367 &10.79 &1426 &41.92 &1608 &47.27 &1 &0.03 \\
Translator19 &3636 &261 &7.18 &1725 &47.44 &1647 &45.30 &3 &0.08 \\
Translator20 &2288 &920 &40.21 &1094 &47.81 &235 &10.27 &39 &1.70 \\
Translator21 &2726 &660 &24.21 &1272 &46.66 &785 &28.80 &9 &0.33 \\
Translator22 &2669 &594 &22.26 &1219 &45.67 &756 &28.33 &100 &3.75 \\
Translator23 &2298 &349 &15.19 &1452 &63.19 &458 &19.93 &39 &1.70 \\
Translator24 &2088 &250 &11.97 &1266 &60.63 &559 &26.77 &13 &0.62 \\
Translator25 &3776 &770 &20.39 &2157 &57.12 &842 &22.30 &7 &0.19 \\
Translator26 &3032 &199 &6.56 &2037 &67.18 &781 &25.76 &15 &0.49 \\
\hline
\textbf{Total} &\textbf{28090} &\textbf{4813} &\multicolumn{1}{r}{\textbf{17.31}} &\textbf{14852} &\textbf{53.44} &\textbf{8148} &\textbf{28.32} &\textbf{277} &\textbf{0.93} \\
\hline
\end{tabular}
}
\caption{Translator-wise final decision counts along with their percentages for the cleaning task}
\label{tab:cc_translatorwise_per}
\end{table*}

\begin{figure*}[!htb]
\centering
\includegraphics[width=\textwidth, keepaspectratio]{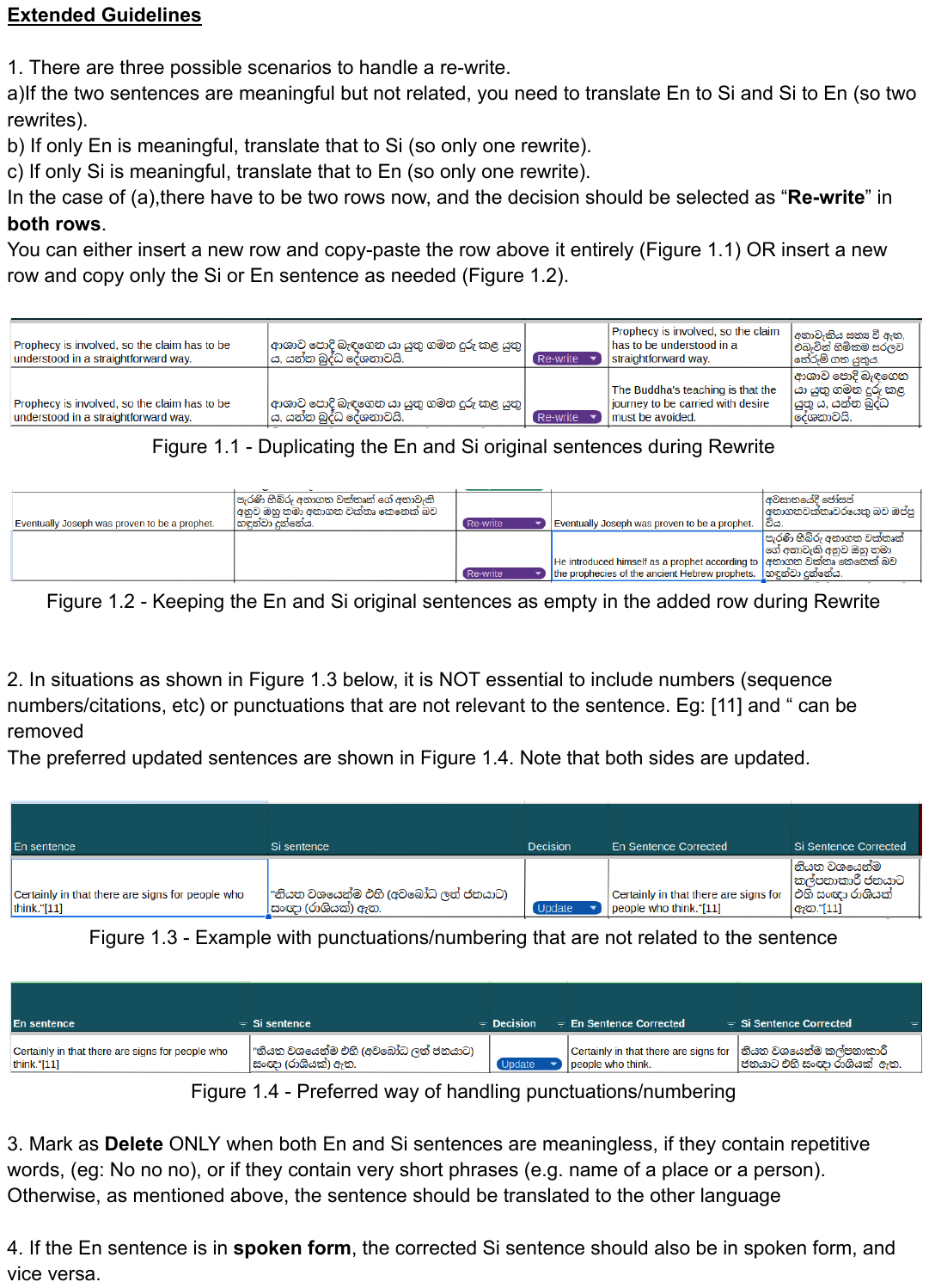}
\caption{Snapshot of the Extended Guidelines given for the translators conducting the web-mining corpus cleaning}
\label{fig:cc_extended_guidelines}
\end{figure*}

Translators were issued a guideline (Figure~\ref{fig:cc_extended_guidelines}) and a demonstration video. The authors reviewed the first 100 sentence pairs cleaned by the translators. Then an Extended Guidelines document was created to cover the common mistakes made during the task and to give specific instructions on the corrective action. The translators were asked to address the reviewer comments given for those hundred sentences. Once the reviewers were satisfied that a translator had fully understood the task, they were given the OK to continue with corpus cleaning. They were asked to record the exact time they spent on the corpus cleaning task. 

Translators were paid as follows: For reading and deciding on the action to be carried out on a sentence pair, a fixed amount was paid. When the translator updates or rewrites sentences, they are paid for each word they write/modify. They were informed of the rates in advance and were given the chance to opt-out of the task after participating in the pilot task.
Table~\ref{tab:cc_translatorwise_per} shows the counts of decisions taken by each translator.

In Table~\ref{tab:cc_translator_durations}, we have summarised the number of sentence pairs cleaned by each translator and the total time taken for it. Owing to the availability of translators, the number of sentence-pairs cleaned by each person was different. Therefore in our calculation, the average time spent by each translator to clean 100 sentence-pairs was considered. Based on these statistics, to clean a sample of hundred sentence pairs from the top 25k of the~\EnSi{} corpus, an average duration of 3hrs 3 minutes with a standard deviation of 1hr and 9 minutes was taken. For~\EnTa{} the average duration was 3hrs 47minutes with a standard deviation of 3hrs 57minutes. We contacted the translators to confirm the times they reported. The three translators who have taken the longest time revealed that they have been recovering from illness/accident, therefore the work had been slow.

\begin{table*}[!ht]\centering
\scriptsize
\resizebox{\textwidth}{!}{%
\begin{tabular}{lrrr|rrrrr}
\hline
\multicolumn{4}{c}{\textbf{En-Si}} &\multicolumn{4}{c}{\textbf{En-Ta}} \\
\hline
\textbf{Translator} &\makecell[c]{\textbf{Total} \\\textbf{Sentences}} &\makecell[c]{\textbf{Duration} \\\textbf{(hh:mm)}} &\makecell[c]{\textbf{Duration for} \\\textbf{100 sentence-pairs}\\\textbf{(hh:mm)}} &\textbf{Translator} &\makecell[c]{\textbf{Total} \\\textbf{Sentences}} &\makecell[c]{\textbf{Duration} \\\textbf{(hh:mm)}} &\makecell[c]{\textbf{Duration for} \\\textbf{100 sentence-pairs}\\\textbf{(hh:mm)}} \\
\hline
Translator16 &652 &30:00 &4:36 &Translator27 &6039 &64:30 &1:04 \\
Translator17 &1523 &19:00 &1:15 &Translator28 &2883 &50:00 &1:44 \\
Translator18 &3402 &146:42 &4:19 &Translator29 &6593 &195:55 &2:58 \\
Translator19 &3636 &97:55 &2:42 &Translator30 &103 &12:00 &11:39 \\
Translator20 &2288 &28:10 &1:14 &Translator31 &105 &7:40 &7:18 \\
Translator21 &2726 &95:00 &3:29 &Translator32 &151 &22:00 &14:34 \\
Translator22 &2669 &38:35 &1:27 &Translator33 &331 &6:03 &1:49 \\
Translator23 &2298 &83:00 &3:37 &Translator34 &102 &6:00 &5:52 \\
Translator24 &2088 &92:00 &4:24 &Translator35 &2784 &39:10 &1:24 \\
Translator25 &3776 &125:31 &3:19 &Translator36 &722 &24:40 &3:24 \\
Translator26 &3032 &97:00 &3:12 &Translator37 &199 &1:30 &0:45 \\
& & & &Translator10 &1707 &27:57 &1:38 \\
& & & &Translator12 &3785 &61:10 &1:36 \\
& & & &Translator38 &457 &9:07 &1:59 \\
& & & &Translator39 &459 &6:10 &1:20 \\
& & & &Translator40 &381 &6:00 &1:34 \\
\hline
\textbf{Totals} &\textbf{28090} &\textbf{853:18} &\textbf{33:36} &\textbf{Totals} &\textbf{26801} &\textbf{539:52} &\textbf{60:44} \\
\textbf{Average (SD)} &\textbf{} &\textbf{} &\textbf{3:03 (1:09)} &\textbf{Average (SD)} &\textbf{} &\textbf{} &\textbf{3:47 (3:57)} \\
\hline
\end{tabular}}
\caption{Cleaning duration analysis for Translators}
\label{tab:cc_translator_durations}
\end{table*}

To compare these durations to what was taken for translating from scratch, we then asked three translators to provide fresh \Si{} translations for a hundred \En{} sentences and to record the time taken. Then we calculated and compared the average time taken along with the standard deviation with the values obtained for the corpus cleaning duration. This information is available in Table~\ref{tab:cc_vs_translation_durations}. The difference between the averages comes to 14 minutes, which means as the number of sentences increases, the time taken for the cleaning task will further be increased.
\begin{table}[!htp]
\resizebox{\linewidth}{!}{%
\begin{tabular}{lrr}
\hline
\multirow{2}{*}{\textbf{}} &\multicolumn{2}{c}{\textbf{Time taken (hh:mm)}} \\
\hline
&\makecell[c]{\textbf{Translation of}\\\textbf{100 sentence-pairs}} &\makecell[c]{\textbf{Cleaning of 100}\\\textbf{sentence-pairs}} \\
\hline
Translator18 &03:06 &4:19 \\
Translator21 &04:12 &3:29 \\
Translator26 &04:25 &3:12 \\
\hline
\textbf{Total Duration} &\textbf{11:43} &\textbf{11:00} \\
\hline
\textbf{Average (SD)} &\textbf{03:54 (00:35)} &\textbf{03:40 (0:28)} \\
\hline
\end{tabular}}
\caption{Time spent to translate 100 En sentences from scratch and for cleaning of 100 \EnSi{} sentence-pairs}
\label{tab:cc_vs_translation_durations}
\end{table}


\section{Model Details}
\label{app:model_details}
As discussed under Section~\ref{sec:NMT_model_performance}, experiments were performed using three state-of-the-art (SOTA) NMT models (NLLB, mBART, and M2M) and vanilla transformer. To perform a fair evaluation between the SOTA NMT models, we chose model variants with similar model sizes. The model sizes utilized in our experiments for NLLB, mBART, and M2M are approximately 600 Million (M), 600M, and 418M, respectively. We perform bilingual fine-tuning on the SOTA models by training up to 3 epochs with a learning rate of $5 \times 10^{-5}$, maximum token length of 200 was set for both source and target. A batch-size of 10 was used for fine-tuning. We utilized the implementations provided by the HuggingFace Transformer library~\cite{wolf-etal-2020-transformers}, and Nvidia Quadro RTX 6000 for hardware-level parallelism. For the decoding process, default settings provided by HuggingFace were retained for each model. In the case of mBART and M2M, a beam search with a beam size of 5 was employed, while for NLLB, a beam size of 4 was utilized.

\paragraph{Vanilla-transformer:}We train the Transformer models implemented in FAIRSEQ library~\cite{ott-etal-2019-fairseq} for our experiments. We train a model consisting of 6 encoder and decoder layers, encoder and decoder embedding of 256, 2 attention heads, dropout of 0.4, the learning rate of $1 \times 10^{-7}$, weight decay of $1 \times 10^{-4}$ and a batch-size of 32. For decoding, we use beam search with a beam size of 5.

\section{Extended Misalignment Analysis}
\label{sec:badparallel}

Table~\ref{tab:BadParallelLong} is an extended version of Table~\ref{tab:BadParallel}. As mentioned in the discussion in Section~\ref{sec:qualitative_analysis}, there are some interesting observations where text from the \textit{Bible} has been aligned with \textit{Buddhist} scripture. There were indeed multiple occurrences of \textit{Bible} being aligned with the \textit{Quran}. However, unlike in the case of \textit{Bible}-\textit{Buddhist} pairings, we are not showing \textit{Bible}-\textit{Quran} pairings here given that both of them being Abrahamic religions~\citep{albayrak2018exploring}, they do share some information and it is reasonable for even a human evaluator to align some of these scripture by mistake. In the last \EnSi{} example, it is also evident that the sentence structure arising from the use of parentheses has played a part in aligning the wrong sentences. In row number 12 of the extended version \EnTa, another interesting observation is that the punctuation count (specifically the quotation marks) has also been a contributor to the misalignment.

\begin{table*}[!htb]
\centering
\includegraphics[width=0.95\textwidth]{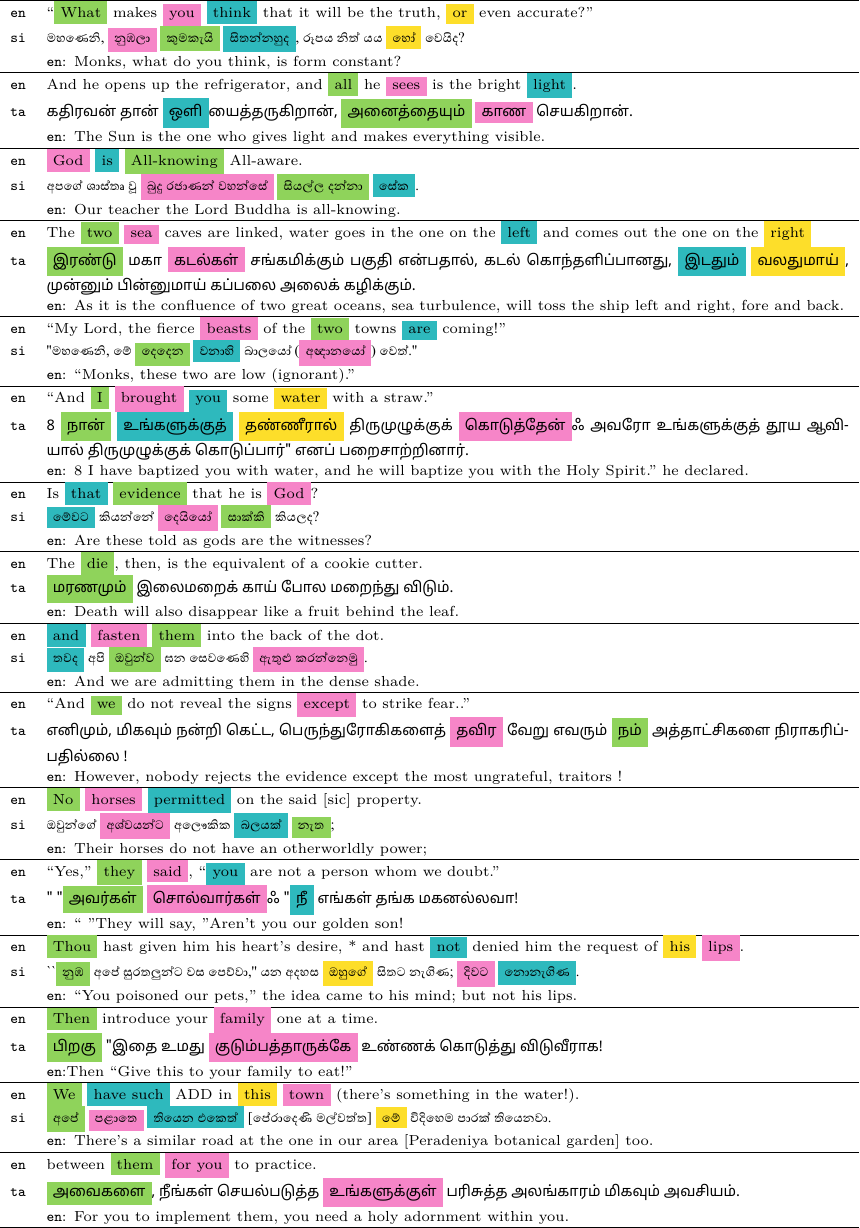}
\caption{Extended set of examples of \textit{parallel} sentences from NLLB where the translated \Si{} or \Ta{} sentence has a different meaning than the original \En{} sentences. We highlighted in colour code the pairs of semantically close words that possibly contributed to the misalignment. Correct \En{} translation of the \Si{} or \Ta{} sentence is given for comparison. }
\label{tab:BadParallelLong}
\end{table*}


\section{NMT Results}
\label{app:nmt_results}
As discussed in Section~\ref{sec:NMT_model_performance} we report the Chrf++ as our primary evaluation metric. Apart from this we also calculate the Chrf, BLEU, and spBLEU scores as well. Since HuggingFace library doesn't support spBLEU score, we are only able to report spBLEU for vanilla-transformer. Tables~\ref{tab:fig1_raw_data}, \ref{tab:fig2_raw_data}, \ref{tab:fig3_raw_data}, \ref{tab:fig4_raw_data}, and \ref{tab:fig5_raw_data} contain the raw results for Figures~\ref{fig:all_datasets_comp_flores_gov}, \ref{fig:models_comp_flores_gov}, \ref{fig:models_comp_flores}, \ref{fig:gov_v_nllb_cleaned}, \ref{fig:gov_v_nllb_enta} and~\ref{fig:labse_v_laser3} respectively.


\begin{table*}[!htb]
    \centering
\resizebox{\textwidth}{!}{%
\tiny
    \begin{tabular}{ll|*{4}{d{2.2}}|*{4}{d{2.2}}}
        \hline
        \multicolumn{2}{l|}{\multirow{2}{*}{Dataset}} & \multicolumn{4}{c|}{SITA}& \multicolumn{4}{c}{FLORES}\\
        \hhline{~~--------}
        & & \EvaluationMetric & \EvaluationMetricEnd \\
        \hline
        \multirow{3}{*}{NLLB} & Top & 17.90 & 15.70 & 2.90 & 0.70 & 22.30 & 20.10 & 5.60 & 1.10 \\ 
        & Random & 8.20 & 6.90 & 0.20 & 0.00 & 7.90 & 6.90 & 0.30 & 0.00\\
        & Bottom & 9.30 & 7.80 & 0.30 & 0.00 & 8.40 & 7.40 & 0.30 & 0.00\\         
        \hline
        \multirow{3}{*}{CCMatrix} & Top & 24.90 & 22.40 & 8.00 & 1.90 & 24.10 & 21.90 & 8.40 & 1.70\\ 
        & Random & 5.60 & 4.60 & 0.10 & 0.00 & 5.70 & 4.80 & 0.10 & 0.00 \\
        & Bottom & 8.70 & 7.10 & 0.10 & 0.00 & 9.80 & 8.00 & 0.00 & 0.00\\         
        \hline
        \multirow{3}{*}{CCAligned} & Top & 26.80 & 24.40 & 10.30 & 2.70 & 22.80 & 21.10 & 8.40 & 1.80 \\ 
        & Random & 23.70 & 20.90 & 7.80 & 1.20 & 19.80 & 17.80 & 4.20 & 0.80\\
        & Bottom & 6.20 & 5.10 & 0.20 & 0.00 & 5.50 & 4.60 & 0.10 & 0.00\\         
        \hline
        \multirow{3}{*}{Wikimatrix} & Top & 21.20 & 18.60 & 5.40 & 0.60 & 23.20 & 20.70 & 7.90 & 1.00\\ 
        & Random & 10.40 & 8.90 & 0.70 & 0.00 & 11.80 & 10.40 & 1.20 & 0.00 \\
        & Bottom & 8.50 & 7.10 & 0.30 & 0.00 & 9.00 & 7.70 & 0.40 & 0.00\\         
        \hline
    \end{tabular}
    }
    \caption{Chrf++ scores visualized in Figure~\ref{fig:all_datasets_comp_flores_gov} as well as other scores used for evaluation.} 
    \label{tab:fig1_raw_data}
\end{table*}
\begin{table*}[!htb]
    \centering
\resizebox{\textwidth}{!}{%
\tiny
    \begin{tabular}{ll|*{4}{d{2.2}}|*{4}{d{2.2}}}
        \hline
        \multicolumn{2}{l|}{\multirow{2}{*}{NMT Model}} & \multicolumn{4}{c|}{SITA}& \multicolumn{4}{c}{FLORES}\\
        \hhline{~~--------}
        & & \EvaluationMetric & \EvaluationMetricEnd \\
        \hline
        \multirow{3}{*}{vanilla-transformer} & Top & 24.90 & 22.40 & 8.00 & 1.90 & 24.10 & 21.90 & 8.40 & 1.70 \\ 
        & Random & 5.60 & 4.60 & 0.10 & 0.00 & 5.70 & 4.80 & 0.10 & 0.00\\
        & Bottom & 8.80 & 7.10 & 0.10 & 0.00 & 9.80 & 8.00 & 0.00 & 0.00\\         
        \hline
        \multirow{3}{*}{mBART} & Top & 41.37 & 37.33 & - & 9.95 & 37.36 & 34.24 & - & 9.08\\ 
        & Random & 31.62 & 28.20 & - & 5.54 & 34.32 & 30.88 & - & 5.89 \\
        & Bottom & 12.12 & 10.24 & - & 0.67 & 16.30 & 14.01 & - & 1.13\\         
        \hline
        \multirow{3}{*}{M2M} & Top & 37.61 & 33.85 & - & 8.23 & 34.76 & 31.78 & - & 8.03 \\ 
        & Random & 25.66 & 22.89 & - & 4.13 & 29.10 & 26.29 & - & 4.75\\
        & Bottom & 10.05 & 8.50 & - & 0.71 & 13.89 & 11.99 & - & 1.02\\         
        \hline
        \multirow{3}{*}{NLLBm} & Top & 47.01 & 42.29 & - & 11.96 & 45.69 & 41.81 & - & 12.73\\ 
        & Random & 45.03 & 40.35 & - & 11.16 & 44.10 & 40.05 & - & 11.43 \\
        & Bottom & 41.89 & 37.36 & - & 8.91 & 42.15 & 38.12 & - & 10.19\\         
        \hline
    \end{tabular}
    }
    \caption{Chrf++ scores visualized in Figure~\ref{fig:models_comp_flores_gov} as well as other scores used for evaluation.}    \label{tab:fig2_raw_data}
\end{table*}

\begin{table*}[!htb]
    \centering
\resizebox{\textwidth}{!}{%
\tiny
    \begin{tabular}{ll|*{4}{d{2.2}}|*{4}{d{2.2}}}
        \hline
        \multicolumn{2}{l|}{\multirow{2}{*}{Dataset Size}} & \multicolumn{4}{c|}{SITA}& \multicolumn{4}{c}{FLORES}\\
        \hhline{~~--------}
        & & \EvaluationMetric & \EvaluationMetricEnd \\
        \hline
        \multirow{1}{*}{0.1 M} & \EnSi & 31.8 & 28.8 & 13.2 & 4.1 & 29.5 & 27.2 & 12.4 & 3.9 \\ 
        \hline
        \multirow{1}{*}{0.2 M} & \EnSi & 34.2 & 30.8 & 15.3 & 4.4 & 32.8 & 30.0 & 15.0 & 4.6 \\ 
        \hline
        \multirow{1}{*}{0.3 M} & \EnSi & 34.2 & 30.8 & 14.1 & 4.3 & 32.2 & 29.5 & 14.1 & 4.4 \\ 
        \hline
        \multirow{1}{*}{0.4 M} & \EnSi & 32.6 & 29.4 & 13.6 & 4.1 & 31.9 & 29.3 & 14.0 & 4.1 \\ 
        \hline
        \multirow{1}{*}{0.5 M} & \EnSi & 32.2 & 29.0 & 13.2 & 3.8 & 31.7 & 29.1 & 13.5 & 4.0 \\ 
        \hline
        \multirow{1}{*}{0.6 M} & \EnSi & 32.2 & 29.0 & 12.5 & 3.7 & 32.2 & 29.5 & 13.4 & 4.3 \\ 
        \hline
        \multirow{1}{*}{0.7 M} & \EnSi & 30.9 & 27.9 & 11.7 & 3.7 & 30.9 & 28.5 & 12.8 & 4.1 \\ 
        \hline
        \multirow{1}{*}{0.8 M} & \EnSi & 29.7 & 26.8 & 10.9 & 3.4 & 30.1 & 27.6 & 12.0 & 3.9 \\ 
        \hline
        \multirow{1}{*}{0.9 M} & \EnSi & 28.5 & 25.8 & 10.3 & 3.3 & 28.7 & 26.4 & 11.3 & 3.6 \\ 
        \hline
        \multirow{1}{*}{1.0 M} & \EnSi & 27.9 & 25.3 & 10.4 & 3.0 & 28.8 & 26.5 & 11.1 & 3.6 \\ 
        \hline
        \multirow{1}{*}{1.1 M} & \EnSi & 28.6 & 25.8 & 10.0 & 3.1 & 29.3 & 26.9 & 10.9 & 3.5 \\ 
        \hline
        \multirow{1}{*}{1.2 M} & \EnSi & 27.6 & 24.9 & 9.7 & 2.9 & 28.8 & 26.4 & 10.8 & 3.6 \\ 
        \hline
        \multirow{1}{*}{1.3 M} & \EnSi & 26.2 & 23.7 & 8.7 & 2.7 & 27.6 & 25.3 & 9.8 & 3.0 \\ 
        \hline
        \multirow{1}{*}{1.4 M} & \EnSi & 27.3 & 24.7 & 9.1 & 2.8 & 28.3 & 25.9 & 10.1 & 3.3 \\ 
        \hline
        \multirow{1}{*}{1.5 M} & \EnSi & 26.4 & 23.8 & 8.6 & 2.6 & 28.2 & 25.9 & 9.9 & 3.3 \\ 
        \hline
        \multirow{1}{*}{1.6 M} & \EnSi & 25.0 & 22.7 & 7.6 & 2.5 & 26.6 & 24.4 & 8.6 & 2.8 \\ 
        \hline
    \end{tabular}
    }
    \caption{Raw values of ChrF++ scores visualized in Figure~\ref{fig:models_comp_flores} as well as other scores used for evaluation.}
    \label{tab:fig3_raw_data}
\end{table*}

\begin{table*}[!htb]
    \centering
\resizebox{\textwidth}{!}{%
    \begin{tabular}{ll|*{4}{d{2.2}}|*{4}{d{2.2}}}
        \hline
        \multicolumn{2}{l|}{\multirow{2}{*}{Dataset}} & \multicolumn{4}{c|}{SITA}& \multicolumn{4}{c}{FLORES}\\
        \hhline{~~--------}
        & & \EvaluationMetric & \EvaluationMetricEnd \\
        \hline
        \multirow{2}{*}{SITA EnSi} & SITA-top25K & 46.20 & 43.10 & 29.00 & 15.30 & 21.70 & 18.80 & 3.70 & 0.40 \\ 
        & SITA-Random25K & 40.80 & 37.90 & 24.00 & 11.30 & 18.90 & 16.30 & 2.10 & 0.20\\   
        \hline
        \multirow{1}{*}{NLLB Original EnSi} & NLLB-Original-Tok25K & 17.90 & 15.70 & 2.90 & 0.70 & 22.30 & 20.10 & 5.60 & 1.10\\   
        \hline
        \multirow{2}{*}{NLLB-Cleaned EnSi} & NLLB-Cleaned-Top25K & 19.20 & 17.00 & 3.70 & 0.80 & 24.30 & 21.90 & 6.60 & 1.70 \\ 
        & NLLB-Cleaned-Complete(27K+) & 19.10 & 16.80 & 3.40 & 0.70 & 23.70 & 21.40 & 6.80 & 1.40\\      
        \hline                
        \multirow{2}{*}{SITA EnTa} & SITA-top25K & 43.80 & 39.00 & 21.50 & 9.10 & 25.00 & 20.90 & 2.00 & 0.20 \\ 
        & SITA-Random25K & 40.50 & 35.80 & 17.70 & 7.30 & 22.40 & 18.50 & 1.10 & 0.00\\   
        \hline
        \multirow{1}{*}{NLLB Original EnTa} & NLLB-Original-Tok25K & 24.30 & 20.50 & 2.60 & 0.40 & 30.40 & 26.20 & 6.30 & 1.10\\   
        \hline
        \multirow{2}{*}{NLLB-Cleaned EnTa} & NLLB-Cleaned-Top25K & 25.20 & 21.40 & 2.90 & 0.70 & 31.50 & 27.20 & 6.60 & 1.10 \\ 
        & NLLB-Cleaned-Complete(26K+) & 26.40 & 22.50 & 3.30 & 0.70 & 32.70 & 28.40 & 7.50 & 1.20\\      
        \hline
    \end{tabular}
    }
    \caption{Chrf++ scores visualized in Figure~\ref{fig:gov_v_nllb_cleaned} and Figure~\ref{fig:gov_v_nllb_enta} as well as other scores used for evaluation.}    \label{tab:fig4_raw_data}
\end{table*}

\begin{table*}[!htb]
    \centering
\resizebox{\textwidth}{!}{%
\tiny
    \begin{tabular}{ll|*{4}{d{2.2}}|*{4}{d{2.2}}}
        \hline
        \multicolumn{2}{l|}{\multirow{2}{*}{NMT Model}} & \multicolumn{4}{c|}{SITA}& \multicolumn{4}{c}{FLORES}\\
        \hhline{~~--------}
        & & \EvaluationMetric & \EvaluationMetricEnd \\
        \hline
        \multirow{3}{*}{LASER-3} & Top & 24.90 & 22.40 & 8.00 & 1.90 & 24.10 & 21.90 & 8.40 & 1.70 \\ 
        & Random & 5.60 & 4.60 & 0.10 & 0.00 & 5.70 & 4.80 & 0.10 & 0.00\\
        & Bottom & 8.80 & 7.10 & 0.10 & 0.00 & 9.80 & 8.00 & 0.00 & 0.00\\         
        \hline
        \multirow{3}{*}{LaBSE} & Top & 9.70 & 8.80 & 0.90 & 0.20 & 10.50 & 9.60 & 0.90 & 0.00\\ 
        & Random & 7.80 & 6.60 & 0.20 & 0.00 & 8.30 & 7.20 & 0.30 & 0.00 \\
        & Bottom & 7.60 & 6.10 & 0.00 & 0.00 & 8.70 & 6.90 & 0.10 & 0.00\\         
        \hline
    \end{tabular}
    }
    \caption{Chrf++ scores visualized in Figure~\ref{fig:labse_v_laser3} as well as other scores used for evaluations.}   
    \label{tab:fig5_raw_data}
\end{table*}

\section{Domain Divergence Evaluation}
\label{app:dom_Div}

We calculate the Jensen Shannon Divergence (JS-div)~\cite{lu-etal-2020-diverging} between the training datasets (NLLB original, NLLB Cleaned top 25K, NLLB Cleand Complete (27K+), SITA Top25K, and SITA Random 25K) and the test sets (SITA and FLORES). We use the code implementation used by~\cite{nayak2023leveraging}. The results of the JS-div can be found in Table~\ref{tab:domain_divergence}.

\begin{table*}[!htp]\centering
\scriptsize
\begin{tabular}{l|d{5.0}|d{5.0}|d{5.0}|d{5.0}|d{5.0}|d{2.2}}
\hline
Datasets & 
\multicolumn{1}{c}{NLLB Top 25K BC} & \multicolumn{1}{|c}{WikiMatrix Top 25K} & \multicolumn{1}{|c}{CCAligned Top 25K} & \multicolumn{1}{|c}{CCMatrix Top 25K} & \multicolumn{1}{|c}{SITA Top 25K} & \multicolumn{1}{|c}{NLLB Top 25K AC} \\ \hline
SITA Test Set & 0.71 & 0.55 & 0.64 & 0.59 & 0.16 & 0.69\\
FLORES Test Set & 0.51 & 0.44 & 0.61 & 0.51 & 0.62 & 0.47\\
\hline
\end{tabular}
\caption{Domain divergence between datasets for \EnSi{}. \texttt{BC}- Before cleaning, \texttt{AC}- after cleaning}
\label{tab:domain_divergence}
\end{table*}

JS-div calculation can be described as follows. It is calculated between two distributions $P$ and $Q$ using the formula shown in Equation~\ref{eq:JSD}, where $M$ is an equally weighted sum of  $M = \frac{1}{2}P + \frac{1}{2}Q$ and $KL(\cdot || \cdot)$ represents the Kullback–Leibler divergence~\cite{kullback1951information}. 

\begin{equation}
    JSD(P||Q) = \frac{1}{2}KL(P||M) + \frac{1}{2}KL(Q||M)
    \label{eq:JSD}
\end{equation}

JS-div ranges from 0 to 1 with lower values indicating that the two distributions are more similar.

We calculate the JS-div for each of the test datasets (SITA and FLORES) against the following portions of the web-mined corpora: NLLB top 25k, WikiMatrix top 25k, CCAligned top 25k, SITA top 25k and NLLB top 25k.


\end{document}